\documentclass[11pt]{article}

\usepackage[preprint]{acl}
\usepackage{times}
\usepackage{latexsym}
\usepackage[T1]{fontenc}
\usepackage[utf8]{inputenc}
\usepackage{microtype}
\usepackage{inconsolata}
\usepackage{graphicx}
\usepackage{amsmath}
\usepackage{amssymb}
\usepackage{booktabs}
\usepackage{tabularx}
\usepackage{array}
\usepackage{multirow}
\usepackage[table]{xcolor}
\usepackage{placeins}
\usepackage{forest}
\usepackage{fontawesome5}
\usepackage{etoc}

\definecolor{placeholderred}{RGB}{190,35,55}
\definecolor{taxsingle}{HTML}{CC79A7}
\definecolor{taxcross}{HTML}{0072B2}
\definecolor{taxadaptive}{HTML}{E69F00}
\definecolor{taxfoundation}{HTML}{009E73}

\newcommand{\yes}{\textcolor{blue!65!black}{$\checkmark$}}
\newcommand{\no}{\textcolor{gray!75}{$\times$}}
\newcommand{\opt}{\textcolor{gray!75}{Opt.}}
\newcommand{\taxhead}[2]{\textcolor{#1}{\textbf{#2}}\enspace}
\newcommand{\taxgroup}[2]{%
  \textcolor{#1!85!black}{\rule[-0.22ex]{2.2pt}{1.65ex}}%
  \hspace{6pt}\textbf{\textcolor{black!88}{#2}}}
\newcommand{\taxleaflabel}[2]{%
  \cellcolor{#1!5}%
  \textcolor{#1!85!black}{\rule[-0.18ex]{1.35pt}{1.35ex}}%
  \hspace{4.5pt}\textbf{\textcolor{black!88}{#2}}}
\newcommand{\taxrowsep}{%
  \arrayrulecolor{black!16}\cmidrule(l{8pt}r{8pt}){1-3}%
  \arrayrulecolor{black}}
\newcommand{\papercite}[2]{#1~\citep{#2}}
\newcommand{\papercitealias}[3]{#1~(\citeauthor{#2}, #3)}

\newcolumntype{Y}{>{\raggedright\arraybackslash}X}
\newcolumntype{L}[1]{>{\raggedright\arraybackslash}p{#1}}
\newcolumntype{M}{>{\raggedright\arraybackslash}p{3.35cm}}
\newcolumntype{D}{>{\raggedright\arraybackslash}p{1.95cm}}
\newcolumntype{V}{>{\raggedright\arraybackslash}p{1.55cm}}
\newcolumntype{S}{>{\raggedright\arraybackslash}p{2.55cm}}
\newcolumntype{F}{>{\raggedright\arraybackslash}p{3.75cm}}
\newcolumntype{G}{>{\raggedright\arraybackslash}p{2.25cm}}
\newcolumntype{W}{>{\raggedright\arraybackslash}p{1.75cm}}

\title{Are We Testing Too Much? A Survey of Benchmark Compression for Efficient Evaluation of Large Language Models}
\author{\textbf{Guoming Ling\textsuperscript{1}}, \textbf{Yupei Lin\textsuperscript{1*}}, \textbf{Zhongzhan Huang\textsuperscript{1}}, \textbf{Junxin Li\textsuperscript{1}}, \textbf{Junjie Liu\textsuperscript{1}}, \\
  \textbf{Shanshan Zhong\textsuperscript{1}}, \textbf{Hefeng Wu\textsuperscript{1}}, \textbf{Liang Lin\textsuperscript{1}} \\
  \textsuperscript{1}Sun Yat-sen University \quad \textsuperscript{*}Co-first author}

\begin{document}
\etocdepthtag{main}
\maketitle

\begin{abstract}

Benchmarks are essential for evaluating and improving large language models (LLMs). Yet many widely used benchmarks are redundant, making evaluation unnecessarily expensive. Benchmark compression lowers this cost by estimating full-benchmark scores and rankings from a partial evaluation. Research in this area has grown rapidly since 2024, but to our knowledge no survey has systematically organized its methods and findings. We review 119 papers to explain how methods choose what to evaluate, recover full-benchmark results, and verify their reliability. From this analysis, we develop a taxonomy covering compression within individual benchmarks, across benchmark suites, and during adaptive evaluation, together with work on redundancy and reliability. We compare their information needs, total costs, reliability checks, and limitations. Methods reduce evaluation costs by selecting representative subsets, using cheaper or reusable information, and testing adaptively, but risk poor coverage, inaccurate estimates, and sequential delays. The resulting taxonomy and practical guide help researchers choose, validate, and deploy methods for more efficient and reliable LLM evaluation.
\end{abstract}

\section{Introduction}

Recently, a growing number of benchmarks have been introduced to evaluate and improve large language models (LLMs) \citep{cobbe2021gsm8k,hendrycks2021mmlu}. Yet many of these benchmarks contain substantial redundancy. Evidence from prior studies on widely used benchmarks such as LongBench, GSM8K, and MMLU shows that fewer than 20\% of the test samples can preserve results close to those of the full set \citep{polo2024tinybenchmarks,huang2025minilongbench}. This redundancy makes LLM evaluation unnecessarily expensive and slows model development. To address this issue, benchmark compression methods aim to reduce evaluation cost while producing results close to those of the full evaluation \citep{polo2024tinybenchmarks,saranathan2025sublime,huang2025minilongbench,vivek2024anchor}.

\begin{figure}[!t]
  \centering
  \includegraphics[width=0.999\columnwidth]{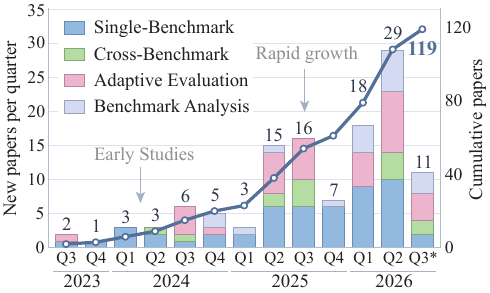}
  \vspace{-20pt}
  \caption{Publication trends in benchmark compression research from 2023 to August 2026. Quarterly and cumulative paper counts show rapid growth since 2024.}
  \label{fig:growth}
  \vspace{-10pt}
\end{figure}

Since 2024, research on benchmark compression has expanded rapidly, as shown in Figure~\ref{fig:growth}. Existing studies approach efficient evaluation in different ways and rely on different information, cost assumptions, and validation methods. These differences make the literature difficult to navigate and its methods difficult to compare. Researchers therefore lack clear guidance on which method fits their setting, which costs it actually reduces, and whether its results remain reliable in practice. 

\begin{figure*}[!t]
  \centering
  \includegraphics[width=0.999\textwidth]{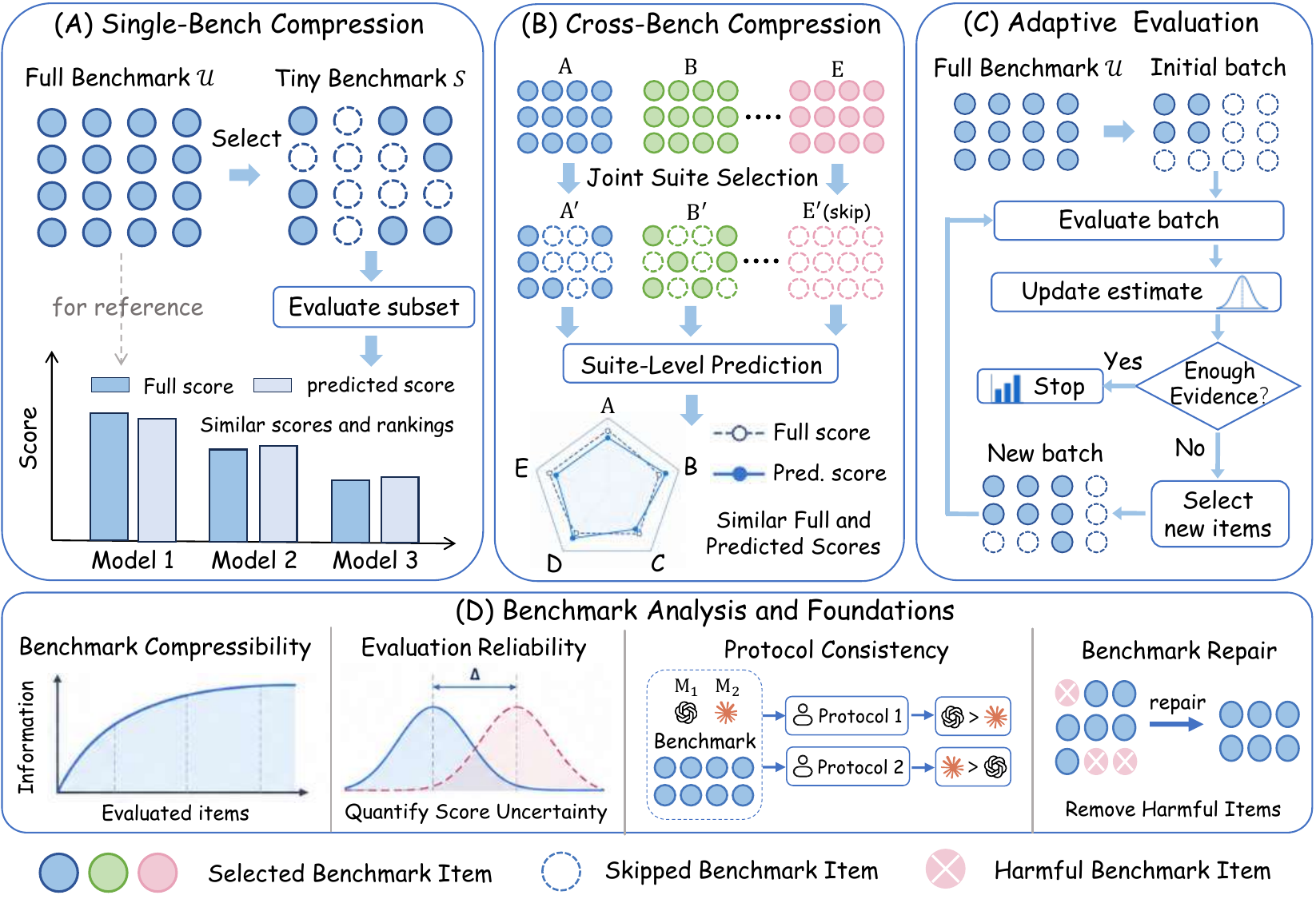}
  \vspace{-20pt}
  \caption{Overall structure of this survey. Sections~\ref{sec:single-benchmark} and~\ref{sec:cross-benchmark} introduce Single-Benchmark Compression and Cross-Benchmark Compression. Sections~\ref{sec:adaptive} and~\ref{sec:foundations} cover Adaptive Evaluation and Benchmark Analysis and Foundations.}
  \label{fig:paradigms}
\end{figure*}

In this survey, we review 119 papers to analyze how benchmark compression methods choose what to evaluate and recover full-benchmark performance from partial evaluations. Based on this analysis, we organize the literature into Single-Benchmark Compression, Cross-Benchmark Compression, Adaptive Evaluation, and Benchmark Analysis and Foundations. We then compare these different types of methods in depth, examining what information they use, what they cost in practice, how they are tested, and where they fail. Across the literature, methods reduce costs through representative subsets, cheaper or reusable information, and adaptive testing. Their main risks are poor coverage, inaccurate estimates, and sequential delays. Finally, we discuss the field's key challenges and future research directions. Together, these analyses provide a foundation for advancing benchmark compression and efficient LLM evaluation. We summarize our contributions as follows:

\begin{itemize}
    \item \textbf{First systematic survey and taxonomy.} To our knowledge, we provide the first systematic survey of LLM benchmark compression, synthesizing 119 papers under a common definition and introducing a principled taxonomy and decision path that organize the field by evaluation setting and available information.
    \item \textbf{Cross-family synthesis and guidance.} We compare method families in depth to clarify their requirements, reliability, and limitations, providing guidance for method selection and validation and a roadmap for future research.
\end{itemize}

\section{Problem Formulation and Taxonomy}
\label{sec:problem}

This section defines the common evaluation contract and then uses it to explain the fixed taxonomy.

\subsection{Problem Formulation}

Let $\mathcal U$ denote the units in a full evaluation. A unit is usually a benchmark item, but may also be a human judgment, an agent run, and so on. A full evaluation obtains an outcome for each unit, yielding $Y_{\mathcal U}$. An aggregation rule $G$ combines these outcomes into the full-evaluation result $\widehat{\theta}_{\mathrm{full}}=G(Y_{\mathcal U})$, such as a score or ranking.
A compressed evaluation obtains outcomes only for a lower-cost subset $\mathcal S\subseteq\mathcal U$, either fixed in advance or chosen adaptively. It may use auxiliary information $Z$, such as item features, historical responses, or a cheap signal from the target model. A recovery rule $R$ uses these partial results to estimate the full-evaluation result as $\widehat{\theta}_{\mathrm{comp}}=R(Y_{\mathcal S},Z)$. The goal is
\[
L(\widehat{\theta}_{\mathrm{comp}},\widehat{\theta}_{\mathrm{full}})\leq\varepsilon
\quad\text{while}\quad
C_{\mathrm{comp}}<C_{\mathrm{full}}.
\]
Here, $L$ measures the relevant difference in score or ranking, and $C$ denotes evaluation cost. In short, benchmark compression uses a lower-cost partial evaluation to approximate the full-evaluation result. Appendix~\ref{app:comparison-formal} gives the complete formulation.

\begin{figure*}[!t]
  \vspace{-1mm}
  \centering
  \definecolor{figthreeblue}{HTML}{4472C4}
  \definecolor{figthreegreen}{HTML}{70AD47}
  \definecolor{figthreepink}{HTML}{E8A2B8}
  \definecolor{figthreefoundation}{HTML}{7B91AE}
  \definecolor{figthreeedge}{HTML}{9DC3E6}
  \resizebox{0.995\textwidth}{!}{%
  \begin{forest}
  for tree={
    grow=east,
    reversed=true,
    anchor=base west,
    parent anchor=east,
    child anchor=west,
    base=center,
    font=\footnotesize,
    rectangle,
    draw=figthreeblue,
    rounded corners=2pt,
    align=center,
    minimum width=4em,
    edge+={figthreeedge, line width=1.0pt, line cap=round, line join=round},
    edge path={\noexpand\path[\forestoption{edge}] (!u.parent anchor) -- ++(7pt,0) |- (.child anchor)\forestoption{edge label};},
    s sep=5pt,
    l sep=9pt,
    inner xsep=4pt,
    inner ysep=2.5pt,
    line width=0.7pt,
    root/.style={rotate=90, child anchor=north, parent anchor=south, anchor=center, font=\small, l sep=12pt},
    category/.style={font=\footnotesize, align=center, minimum width=8.2em, inner xsep=3pt, inner ysep=2pt, l sep=12pt},
    cat-single/.style={category, draw=figthreeblue, fill=figthreeblue!10},
    cat-cross/.style={category, draw=figthreegreen, fill=figthreegreen!10},
    cat-adaptive/.style={category, draw=figthreepink, fill=figthreepink!10},
    cat-foundation/.style={category, draw=figthreefoundation, fill=figthreefoundation!10},
    taxleaf/.style={align=left, font=\footnotesize},
    leaf-single/.style={taxleaf, draw=figthreeblue, fill=figthreeblue!7},
    leaf-cross/.style={taxleaf, draw=figthreegreen, fill=figthreegreen!7},
    leaf-adaptive/.style={taxleaf, draw=figthreepink, fill=figthreepink!7},
    leaf-foundation/.style={taxleaf, draw=figthreefoundation, fill=figthreefoundation!7}
  }
  [Benchmark Compression, root
    [Single-Benchmark\\Compression, cat-single
      [{\parbox[t]{60em}{\taxhead{figthreeblue!85!black}{Content-Based Compression}\papercite{Scales++}{bean2025scales++}; \papercite{PromptEval}{polo2024efficient}; \papercite{EffiEval}{wang2025effieval}; \papercite{MIS Select}{kjorvezir2026consistent}; \papercite{Agreement}{purificato2026select}; \papercite{PRECISE}{divekar2026statistically}; \papercite{CMLE}{shen2026robust}; \papercite{NoisyHT}{feng2026noisy}; \papercite{ARES}{saad2024ares}; \papercite{CoT-Core}{pan2026cot}.}}, leaf-single]
      [{\parbox[t]{60em}{\taxhead{figthreeblue!85!black}{Historical-Response Compression}\papercite{AnchorPoints}{vivek2024anchor}; \papercite{tinyBenchmarks}{polo2024tinybenchmarks}; \papercite{SubLIME}{saranathan2025sublime}; \papercite{MiniLongBench}{huang2025minilongbench}; \papercite{DataLite}{wang2026rethinking}; \papercite{SparseEval}{zhang2026sparseeval}; \papercite{REP-CORE}{zhang2026learning}; \papercite{SMART}{gupta2025improving}; \papercite{AIPW}{zhang2026benchmark}; \papercite{MINCE}{das2026mince}; \papercite{DISCO}{rubinstein2026disco}; \papercite{mRMR++}{bowyer2026efficient}; \papercite{Option IRT}{fei2026every}; \papercite{CollabEval}{fisch2026collabeval}; \papercite{BiasBound}{zhuang2025efficient}; \papercite{ShortForm}{gignac2025psychometrically}; \papercite{LURE}{berrada2026scaling}; \papercite{MetaEval}{wang2026metaeval}; \papercite{Neyman}{liu2026active}; \papercite{DatBench}{joshi2026datbench}; \papercite{SoftElo}{kargi2026uncertain}; \papercite{MidRange}{ndzomga2026efficient}; \papercite{CJE}{landesberg2025causal}; \papercite{BISS}{matricon2025efficiently}; \papercite{JBDistill}{zhang2025jailbreak}; \papercite{Optimal Sampling}{li2025towards}.}}, leaf-single]
      [{\parbox[t]{60em}{\taxhead{figthreeblue!85!black}{Target-Conditioned Compression}\papercite{TAILOREDBENCH}{yuan2025beyond}; \papercite{GAT}{ramakrishnan2026generative}; \papercite{MAD}{feng2025sample}; \papercite{MetricMatch}{unell2026metric}; \papercite{Judge Sampling}{unell2025smarter}; \papercite{ReviewAug}{kim2026augmenting}; \papercite{JudgeAdjust}{lee2025correctly}; \papercite{RidgePPI}{eyre2024regression}; \papercite{QuEst}{deng2025quest}; \papercite{GLIDE}{martinon2026industrializing}; \papercite{CostEval}{angelopoulos2025cost}; \papercite{ControlEval}{zhou2025accelerating}; \papercite{AutoEval}{boyeau2024autoeval}; \papercite{SSME}{shanmugam2026evaluating}; \papercite{DiffUse}{tahan2024label}.}}, leaf-single]
    ]
    [Cross-Benchmark\\Compression, cat-cross
      [{\parbox[t]{60em}{\papercite{metabench}{kipnis2025metabench}; \papercite{BenchPress}{zeng2026you}; \papercite{Adaptive Sampling}{xu2024data}; \papercite{LEGO-IRT}{liao2025toward}; \papercite{FPC}{habba2026growing}; \papercite{Facility}{yao2026coresets}; \papercite{Agent PACE}{song2026pace}; \papercite{Submodular}{smola2026submodular}; \papercite{SimBA}{subramani2025simba}; \papercite{SkillFactors}{maimon2026benchmarks}; \papercite{ResampledBench}{suzuki2025resampling}; \papercite{BIS}{shi2025importance}; \papercite{PSN-IRT}{zhou2026lost}; \papercite{MultiIRT}{lior2026extending}.}}, leaf-cross]
    ]
    [Adaptive\\Evaluation, cat-adaptive
      [{\parbox[t]{60em}{\taxhead{figthreepink!65!black}{Adaptive Item Selection}
        \papercite{AEA}{li2024active}; \papercite{Fluid}{hofmann2025fluid}; \papercite{ATLAS}{li2025adaptive};
        \papercitealias{FactorCAT}{truongbeyond}{2025}; \papercite{V-Optimal}{krumdick2026cost}; \papercite{ReEval}{truong2025reliable};
        \papercite{M3IRT}{uebayashi2026evaluating}; \papercite{BayesAME}{cordero2026bayesame}; \papercite{IRSL}{truong2026item};
        \papercite{AcTracer}{huang2026actracer}; \papercite{AutoJudger}{ding2026autojudger}; \papercite{Cer-Eval}{wang2025cer};
        \papercite{FAQ-PAI}{wu2026efficient}; \papercite{CEM}{kim2026measuring}; \papercite{RIPr}{hsu2026efficient};
        \papercite{IRT-Agent}{ge2026agent}; \papercite{APOT}{asgari2025adaptive}; \papercite{SafetyCAT}{spagliardi2026efficient};
        \papercite{CollabFilter}{zhong2025efficient}; \papercite{ROBIN-HOOD}{saha2026llm};
        \papercite{FAIR-Eval}{anonymous2026faireval}.}}, leaf-adaptive]
      [{\parbox[t]{60em}{\taxhead{figthreepink!65!black}{Sequential Decision and Stopping}\papercite{Flash-HELM}{perlitz2024efficient}; \papercite{UCB-E}{zhou2025speeding}; \papercite{AdaptiveRank}{balkir2026confident}; \papercite{JustEnough}{arviv2026stop}; \papercite{SySRs}{lyu2026cutting}; \papercite{SELECT-LLM}{durmazkeser2026large}; \papercite{CELEUS}{zhou2026celeus}; \papercite{ParEvalLayer}{huang2026parevallayer}; \papercite{PULSE}{tolochinsky2026valid}; \papercite{TrustEscalate}{jung2025trust}; \papercite{AV-AIVAT}{li2026av}; \papercite{R-AutoEval+}{park2026adaptive}; \papercite{HiddenGems}{kahana2026discovering}; \papercite{Arena-Lite}{son2025arena}; \papercite{De-Arena}{yin2026decentralized}.}}, leaf-adaptive]
    ]
    [Analysis and\\Foundations, cat-foundation
      [{\parbox[t]{60em}{\papercite{ONEBench}{ghosh2025onebench}; \papercite{DualEval}{li2026dualeval}; \papercite{Redundancy}{zhang2025redundancy}; \papercite{IRT Audit}{jiang2026can}; \papercite{TaskBudget}{huang2026many}; \papercite{cBMM}{qu2026interpretable}; \papercite{JudgeCal}{li2026calibrate}; \papercite{JudgeDiag}{gupta2026diagnosing}; \papercite{HarnessAudit}{zhang2026stop}; \papercite{BestN}{landesberg2026llm}; \papercite{BenchScope}{sha2026benchscope}; \papercite{OPD}{cao2026does}; \papercite{LaplaceIRT}{francisco2026laplace}; \papercite{MDAD}{yauney2026reliable}; \papercite{ReliableEval}{lior2025reliableeval}; \papercite{Resolution}{kotawala2026resolution}; \papercite{ErrorBars}{miller2024adding}; \papercite{BENCHMARK2}{qian2026benchmark}.}}, leaf-foundation]
    ]
  ]
  \end{forest}%
  }
  \caption{Taxonomy of LLM benchmark compression research. Studies are grouped by their primary contribution. Appendix~\ref{app:taxonomy} explains classification rules and boundary cases. Appendix~\ref{app:catalogue} provides detailed paper tables by category.}
  \label{fig:taxonomy}
\end{figure*}
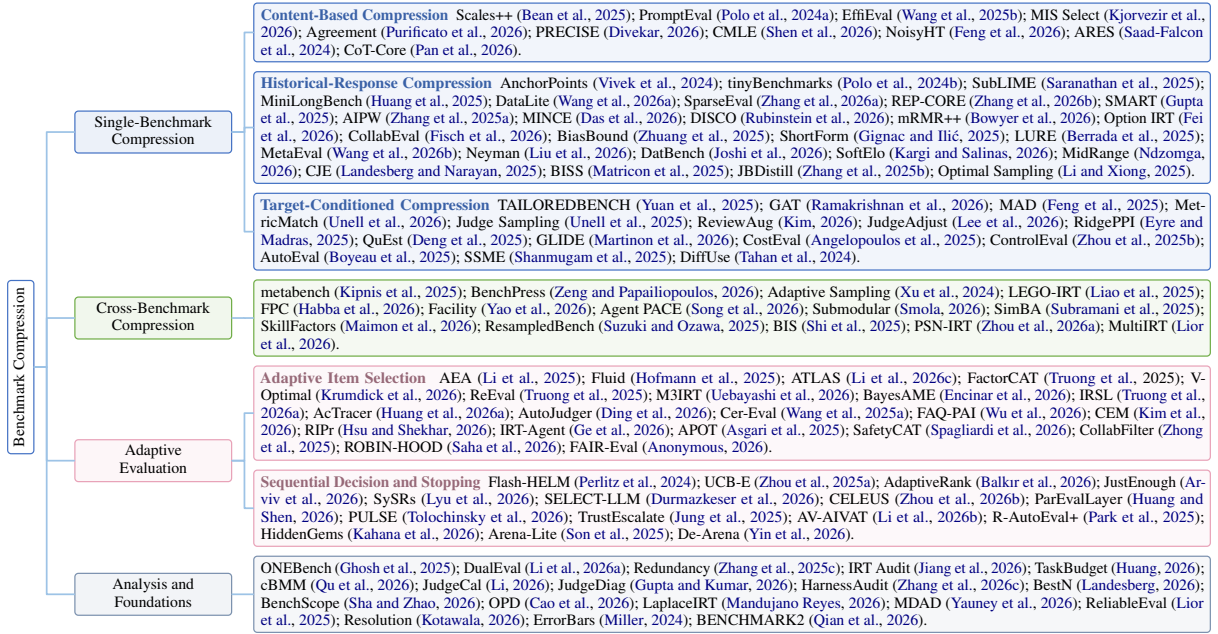

\subsection{Taxonomy Overview}

We classify each paper by asking three questions in order. \textbf{Contribution type.} Is the paper's primary contribution an operational cost-reduction method or foundational analysis of benchmark evaluation and compression? Papers in the latter group form Benchmark Analysis and Foundations. \textbf{Subset selection.} Is the evaluated subset fixed before evaluation begins, or is it selected dynamically during evaluation? The former is static benchmark compression, and the latter is Adaptive Evaluation. Adaptive methods divide into Adaptive Item Selection, which chooses the next item, and Sequential Decision and Stopping, which governs the terminal action. \textbf{Scope and information.} A static method is Cross-Benchmark Compression when it jointly reduces several benchmark evaluations and Single-Benchmark Compression otherwise. Single-Benchmark Compression has three subtypes based on the information used to select items. Content-Based Compression uses benchmark-side information, Historical-Response Compression adds responses from earlier models, and Target-Conditioned Compression adds a cheap signal from the model being evaluated. Because the taxonomy is hierarchical, fixed versus dynamic subset selection is considered before benchmark scope. Scope distinguishes only methods whose evaluated subsets are fixed in advance. Each paper therefore receives one primary label based on its main contribution and evaluation design, while other properties are recorded as secondary attributes. This hierarchy provides a consistent framework for navigating the literature and comparing methods. Figure~\ref{fig:paradigms} and~\ref{fig:taxonomy} show the survey framework and taxonomy, respectively. Appendix~\ref{app:taxonomy} gives the formal definitions and classification rules, and Appendix~\ref{app:taxonomy-hybrids} covers hybrid and boundary cases. The following sections review the work in each category in detail.

\section{Static Benchmark Compression}
\label{sec:static}

\begin{figure}[!t]
  \centering
  \includegraphics[width=0.999\columnwidth]{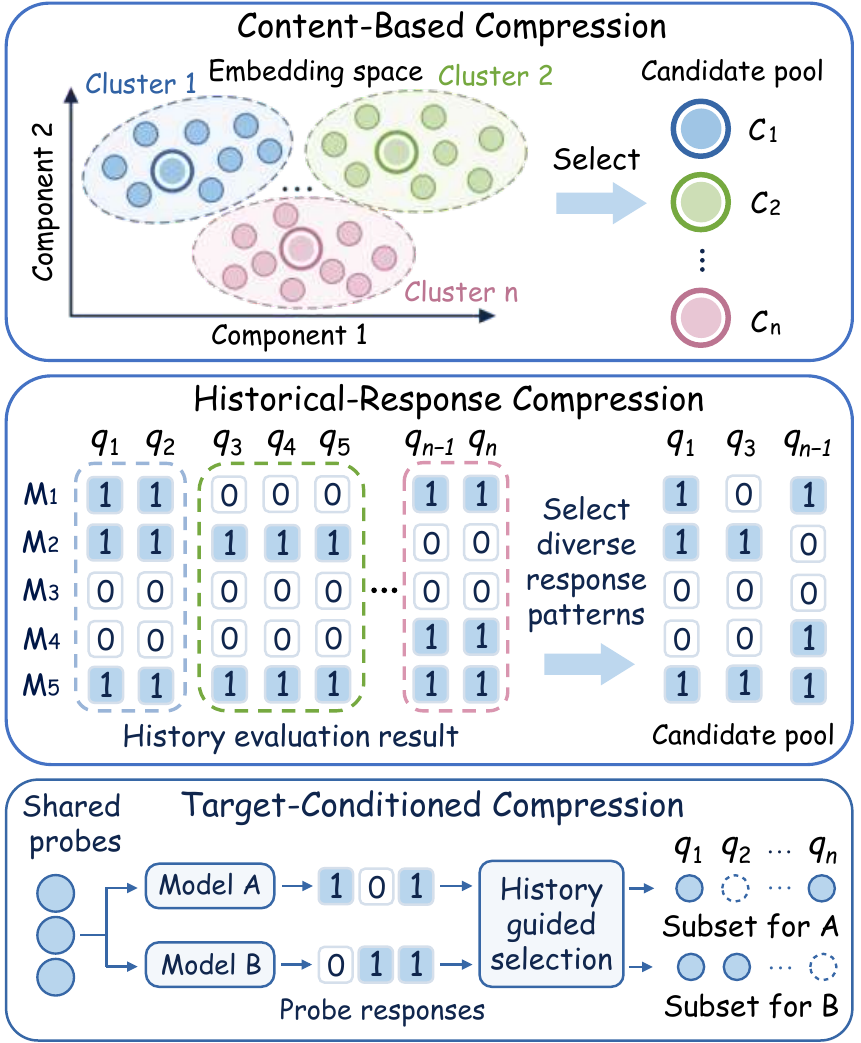}
  \caption{Illustration of single benchmark compression. Subsets are selected using item content, historical responses, or low-cost probes of the evaluated model.}
  \label{fig:single}
\end{figure}

Static Benchmark Compression fixes the evaluation subset before the main evaluation begins. Methods that reduce one benchmark form Single-Benchmark Compression, as illustrated in Figure~\ref{fig:single}, whereas methods that reduce several benchmarks together form Cross-Benchmark Compression.

\subsection{Single-Benchmark Compression}
\label{sec:single-benchmark}

Single-Benchmark Compression evaluates a model on a fixed subset of one benchmark and uses the subset results to estimate its full-benchmark score or ranking. Based on the information used to select the subset, we divide these methods into three groups. Content-Based Compression uses sampling rules or item data, Historical-Response Compression uses results from previously evaluated models, and Target-Conditioned Compression uses a cheap signal from the current model.
Table~\ref{tab:single-representatives} compares representative methods across groups.
The following subsections discuss each group further.

\subsubsection{Content-Based Compression}

Content-Based Compression builds the subset from a sampling rule or item information  such as text, metadata, and representations. 

Content-Based Compression ranges from simple sampling to content-guided selection. Random and stratified sampling use sample means or group weights. PromptEval extends structured sampling to sparse prompt-by-example cells and predicts the missing cells \citep{polo2024efficient}. Coreset methods group items by text, required skills, or reasoning, then aggregate selected representatives directly, with cluster weights, or with predictions for untested items \citep{kjorvezir2026consistent,bean2025scales++,wang2025effieval,pan2026cot}. Auxiliary scores provide another route. Content-guided probability sampling favors informative items and corrects selection with inverse-probability weighting, while calibrated-judge methods use a small human-labeled subset to correct many automated judgments \citep{purificato2026select,saad2024ares,divekar2026statistically,feng2026noisy}. None of these methods requires a historical model-by-item response matrix.

Content-Based Compression has low construction cost. Random and stratified sampling require little setup cost, while representation-based methods avoid collecting a large and expensive model-by-item response matrix. However, small random subsets can have high variance, and items that appear diverse in text or embedding space may still test similar abilities. Using less information can therefore make full-benchmark recovery less stable or accurate. At the same time, weaker reliance on historical response patterns may improve robustness when model behavior shifts.

\subsubsection{Historical-Response Compression}

Historical-Response Compression reuses past evaluation results to reduce the cost of testing new models. These results form a model-by-item response matrix that records each model's performance on each item. Methods use these patterns to select representative or informative items.

Historical responses can guide item coverage, learned recovery, or statistical correction. Coverage and psychometric methods cluster items by response patterns or fit item response theory models, then recover the full score with cluster weights or an estimated ability \citep{vivek2024anchor,polo2024tinybenchmarks,huang2025minilongbench,wang2026metaeval}. Learned-recovery methods treat selected responses as features and predict the full score or missing item results \citep{saranathan2025sublime,zhang2026sparseeval,zhang2026learning,fisch2026collabeval}. Model-assisted methods retain randomized target sampling and use residual or inverse-probability correction to keep the estimate tied to the original item pool \citep{zhang2026benchmark,fisch2026collabeval,berrada2026scaling,liu2026active}. Historical filtering is a boundary case. It keeps difficult or discriminative items and scores the refined benchmark directly rather than reconstructing the original full score \citep{gupta2025improving,joshi2026datbench,zhang2025jailbreak}. Each approach relies on historical patterns transferring to the new model.

Historical results may support smaller subsets or more accurate full-score estimates than item data alone. However, building the response matrix requires many full evaluations unless public results are available. Moreover, under large model shifts, such as substantial performance gains, their advantage may shrink and random sampling can remain competitive \citep{zhang2026benchmark}. Broad historical coverage is therefore important for generalization. Target-Conditioned Compression reduces this dependence by using signals from the current model.

\begin{table*}[!t]
  \centering
  \begingroup
  \fontsize{8.4}{9.8}\selectfont
  \setlength{\tabcolsep}{2.2pt}
  \renewcommand{\arraystretch}{1.08}
  \begin{tabular*}{\textwidth}{@{\extracolsep{\fill}} c l l c c c l l @{\hspace{1pt}}}
    \toprule
    \multirow{2}{*}{\textbf{Type}} & \multirow{2}{*}{\textbf{Method}} & \multirow{2}{*}{\textbf{Domain}} & \multicolumn{3}{c}{\textbf{Information}} & \multirow{2}{*}{\textbf{Selection}} & \multirow{2}{*}{\textbf{Estimation}} \\
    \cmidrule(lr){4-6}
    & & & Features & Responses & Target signal & & \\
    \midrule
    \multirow{3}{*}{\textbf{Content}} & PromptEval~\cite{polo2024efficient} & General & \yes & \no & \no & Balanced & Prediction \\
     & Scales++~\cite{bean2025scales++} & General & \yes & \no & \no & Clustering & Weighted + Pred. \\
     & EffiEval~\cite{wang2025effieval} & General & \yes & \no & \no & Coverage & Mean \\
    \midrule
    \multirow{4}{*}{\textbf{History}} & Anchor Points~\cite{vivek2024anchor} & General & \no & \yes & \no & Clustering & Weighted \\
     & tinyBenchmarks~\cite{polo2024tinybenchmarks} & General & \no & \yes & \no & Clustering & Weighted + Pred. \\
     & MiniLongBench~\cite{huang2025minilongbench} & Long-context & \yes & \yes & \no & Clustering & Prediction \\
     & SparseEval~\cite{zhang2026sparseeval} & General & \no & \yes & \no & Optimization & Prediction \\
    \midrule
    \multirow{3}{*}{\textbf{Target}} & TailoredBench~\cite{yuan2025beyond} & General & \no & \yes & \yes & Clustering & Prediction \\
     & AutoEval~\cite{boyeau2024autoeval} & General & \no & \no & \yes & Sampling & Pred. + Corr. \\
     & CVE~\cite{zhou2025accelerating} & Chat & \no & \no & \yes & Sampling & Pred. + Corr. \\
    \bottomrule
  \end{tabular*}
  \endgroup
  \caption{Comparison of representative single benchmark compression methods. Pred.\ and Corr.\ denote prediction and residual correction. See Appendix~\ref{app:catalogue} for the complete catalogue and Appendix~\ref{app:comparison} for detailed comparisons.}
  \label{tab:single-representatives}
\end{table*}

\subsubsection{Target-Conditioned Compression}

Target-Conditioned Compression uses a low-cost signal from the current model, such as probe results or confidence scores, to select its subset before the main evaluation. Different models can therefore receive different fixed subsets.

The current-model signal may guide a probe, direct item selection, or proxy correction. Probe-based methods evaluate a small shared set, identify similar historical models, and use them to build a tailored subset and predict untested items \citep{yuan2025beyond}. Uncertainty- and disagreement-based methods favor items on which the current model is unsure or the candidates differ. Weighted sampling recovers the full score, while selected pairwise comparisons produce a winner or ranking \citep{ramakrishnan2026generative,feng2025sample,tahan2024label}. Proxy-correction methods score the full pool with a cheap judge and send a small fixed subset to humans. The human labels audit the judge or correct the final estimate, sometimes with more reviews assigned to likely judge errors \citep{unell2026metric,unell2025smarter,kim2026augmenting,angelopoulos2025cost,boyeau2024autoeval}.

Target-Conditioned Compression tailors the subset to the current model while retaining parallel batch execution. It is useful only when the probe or proxy is much cheaper than the tests or reviews saved, so its full cost must include output scans and cheap judge calls. Weak signals can miss important items or introduce systematic proxy errors. Random coverage and paired human checks provide safeguards. Judge-based methods often save human or strong-judge reviews rather than model inference, so the saved unit should be stated clearly.

\subsection{Cross-Benchmark Compression}
\label{sec:cross-benchmark}

Single benchmark compression reduces redundancy within a single benchmark. Benchmark suites can also be redundant because different benchmarks often test similar abilities. Cross-Benchmark Compression selects a smaller set from across the suite and uses its results to estimate the full set of benchmark scores or the suite ranking.

Cross-Benchmark Compression can remove items, complete benchmarks, or expensive target tasks. Suite-anchor methods select informative items across benchmarks and use shared ability models to recover individual scores, a suite average, or a ranking \citep{kipnis2025metabench,liao2025toward,habba2026growing}. Benchmark-selection methods run only a few complete tasks and use score-matrix completion to predict the skipped benchmark scores \citep{zeng2026you,subramani2025simba,maimon2026benchmarks}. Source-to-target methods replace an expensive benchmark with cheaper proxy tasks, then estimate the target score or ranking through regression or importance weighting \citep{song2026pace,shi2025importance}. Multilingual tensor models extend suite-level sharing across languages \citep{lior2026extending}. The saved unit thus grows from individual items to entire benchmark runs or costly environments.

Cross-Benchmark Compression is most useful when one model needs to be evaluated on several benchmarks. Joint compression reduces redundancy both within and across benchmarks, making it more efficient than compressing each benchmark independently. Some methods can even skip complete benchmarks or costly tasks. This sharing also creates the main risk: estimates rely on cross-benchmark relations that may not hold for new models, especially when learned from past suite results. Validation should therefore include unseen models, while unreliable components should be run directly. The next section turns to Adaptive Evaluation, where the subset is not fixed in advance.

\FloatBarrier

\begin{table*}[!t]
  \centering
  \begingroup
  \fontsize{8}{9.8}\selectfont
  \setlength{\tabcolsep}{2.6pt}
  \renewcommand{\arraystretch}{1.08}
  \begin{tabular*}{\textwidth}{@{\extracolsep{\fill}} c l l c c c c c l l @{\hspace{1pt}}}
    \toprule
    \multirow{2}{*}{\textbf{Type}} & \multirow{2}{*}{\textbf{Method}} & \multirow{2}{*}{\textbf{Domain}} & \multicolumn{3}{c}{\textbf{Information}} & \multicolumn{2}{c}{\textbf{Online control}} & \multirow{2}{*}{\textbf{Selection}} & \multirow{2}{*}{\textbf{Estimation}} \\
    \cmidrule(lr){4-6}\cmidrule(lr){7-8}
    & & & Features & Responses & Target signal & Update & Stop & & \\
    \midrule
    \multirow{5}{*}{\textbf{Selection}} & Fluid~\cite{hofmann2025fluid} & General & \no & \yes & \no & \yes & \no & Fisher info. & IRT ability \\
     & AEA~\cite{li2024active} & General & \yes & \yes & \no & \yes & \no & RL & Prediction \\
     & AutoJudger~\cite{ding2026autojudger} & Multimodal & \yes & \yes & \no & \yes & \no & Agent & IRT ability \\
     & FAQ-PAI~\cite{wu2026efficient} & General & \no & \yes & \no & \yes & \no & Uncertainty & Pred. + Corr. \\
     & IRT-Agent~\cite{ge2026agent} & Code agents & \no & \yes & \no & \yes & \no & Fisher info. & IRT ability \\
    \midrule
    \multirow{3}{*}{\textbf{Stopping}} & Flash-HELM~\cite{perlitz2024efficient} & General & \no & \yes & \no & \no & \yes & Sampling & Mean \\
     & CELEUS~\cite{zhou2026celeus} & General & \no & \no & \yes & \no & \yes & Uncertainty & Pred. + Corr. \\
     & R-AutoEval+~\cite{park2026adaptive} & General & \no & \no & \yes & \no & \yes & Sampling & Pred. + Corr. \\
    \bottomrule
  \end{tabular*}
  \endgroup
  \caption{Comparison of representative adaptive evaluation methods. Update and Stop denote feedback-driven selection and stopping. See Appendix~\ref{app:catalogue} for the complete catalogue and Appendix~\ref{app:comparison} for detailed comparisons.}
  \label{tab:adaptive-representatives}
\end{table*}

\section{Adaptive Evaluation}
\label{sec:adaptive}

Adaptive Evaluation uses results collected so far to decide what to test next and when to stop. Adaptive Item Selection chooses the next item to evaluate, while Sequential Decision and Stopping decides whether enough items have been tested. 
Table~\ref{tab:adaptive-representatives} compares representative adaptive methods.

\subsection{Adaptive Item Selection}

Adaptive Item Selection starts with a small set of items. After each response or batch, it updates the score, ability estimate, or uncertainty, then selects the item expected to improve the estimate most.

Adaptive Item Selection chooses the next test through ability estimation, response prediction, or uncertainty-based allocation. Computerized adaptive testing updates an item response theory ability estimate after each answer, selects the most informative next item, and maps the final ability to a score or ranking \citep{hofmann2025fluid,li2025adaptive,anonymous2026faireval}. Active matrix completion predicts the target's untested responses and selects the item expected to improve those predictions most \citep{li2024active,cordero2026bayesame,zhong2025efficient}. Other methods direct more tests to uncertain groups, rare failures, or noisy judge cells. Group weights or sampling probabilities correct uneven selection, while repeated-cell means estimate noisy comparisons \citep{wang2025cer,huang2026actracer,wu2026efficient,kim2026measuring,saha2026llm}. Unlike static compression, each new result can redirect later tests.

Adaptive Item Selection can reduce cost by directing later tests to the items that are most informative for the current model. This flexibility has two main costs. Sequential rounds increase wall-clock time and give models different test paths, which complicates direct comparison and reproduction. Adaptive subsets may also no longer represent the full benchmark, so reliable scores depend on suitable correction and, for some methods, well-calibrated historical data. Complete test-path logs are therefore important. Better item selection still does not determine when evaluation should stop, which is the focus of the next subsection.

\subsection{Sequential Decision and Stopping}

Sequential Decision and Stopping evaluates items in stages and checks after each stage whether more testing is needed. It may target a score with a requested precision, a reliable pairwise comparison, or the best model under a fixed budget. It may also accept, reject, escalate, or abstain.

Sequential methods save cost by stopping at a target precision, focusing a fixed budget on likely winners, or avoiding unnecessary comparisons. Staged methods stop when a score interval, rank, or pairwise difference reaches the required level, while anytime-valid methods allow checking after every result \citep{perlitz2024efficient,balkir2026confident,arviv2026stop,zhou2026celeus,li2026av}. Best-model bandits concentrate tests on plausible winners and eliminate weak candidates, sometimes using shared queries or matrix predictions \citep{zhou2025speeding,lyu2026cutting,tolochinsky2026valid}. Selective-decision methods route easy cases to cheap judges, escalate uncertain ones, or use partial tasks and tournaments to return a promotion decision, abstention, or ranking \citep{huang2026parevallayer,jung2025trust,son2025arena}.

The main advantage of sequential methods is that they spend evaluation effort where it can still change the result. They can stop once a score is precise enough or a comparison is settled, while fixed-budget methods can focus tests on competitive models. Their savings, however, depend on difficulty. Close models may require almost the full benchmark, and fixed-budget methods redistribute rather than reduce the budget. They may also return only a winner or decision instead of full scores. Reliable early stopping requires a valid sequential rule. Staged execution can also increase wall-clock time and requires complete logs for reproduction.

\FloatBarrier

\section{Benchmark Analysis and Foundations}
\label{sec:foundations}

Having reviewed the main benchmark compression methods, we now turn to foundational studies that examine four questions underlying these methods.

\paragraph{Why Can Benchmarks Be Compressed?}
Many benchmarks can be compressed because their items measure overlapping capabilities and provide limited additional discrimination among models. In a model-by-item response matrix, this redundancy appears as correlated model responses, low item discrimination, and low effective dimensionality \citep{ghosh2025onebench,zhang2025redundancy,qu2026interpretable,sha2026benchscope}. This structure allows smaller subsets to retain information needed for score or ranking recovery. However, redundancy depends on the evaluated models and the target result. A subset that preserves an overall ranking may still distort score gaps or reverse close models.

\paragraph{How Many Items Can Be Removed?}
There is no universally safe compression ratio across benchmarks. Item response models need suitable calibration models and stable item parameters, whose uncertainty can affect selection and model comparisons \citep{li2026dualeval,jiang2026can,francisco2026laplace}. The required item count also depends on the score difference of interest, paired variance, task-group coverage, and allowed error rate \citep{huang2026many,yauney2026reliable,kotawala2026resolution,miller2024adding}. Random sampling can remain competitive with more complex selectors. When retained items cannot reliably separate two models, their comparison should remain unresolved.

\paragraph{Which Settings Must Stay Fixed?}
Full and compressed evaluations should use the same judges, prompts, decoding settings, and agent harnesses. Aggregate judge accuracy can hide local errors, and removing weaker judges can discard useful signals \citep{li2026calibrate,gupta2026diagnosing,landesberg2026llm}. Prompt and harness changes can also alter model comparisons \citep{lior2025reliableeval,zhang2026stop}. Any protocol change should therefore be assessed separately from compression.

\paragraph{Compression or Benchmark Repair?}
Not every smaller benchmark is intended to reproduce the original result. Some methods remove questions whose answers can be guessed from option wording or whose outcomes conflict with the intended capability ordering \citep{cao2026does,qian2026benchmark}. Their goal is to build a cleaner benchmark, even if model scores or rankings change. This is benchmark repair rather than score reconstruction. Compression asks whether fewer evaluations can reproduce the original result, while repair asks whether that result should remain the target. A repaired benchmark may be smaller and cheaper, but it must be validated as a new benchmark.

\section{Comparison Across Method Families}
\label{sec:comparison}

We next compare the six compression families to guide method choice. We focus on subset selection, score recovery, lifecycle cost and limitations. Appendix~\ref{app:comparison} gives the detailed comparison.

\subsection{Subset Selection: Information and Timing}

Content-Based and Historical-Response Compression select a fixed subset before testing the target model. The former uses benchmark data, while the latter uses responses from earlier models. 
Target-Conditioned Compression first collects a cheap signal from the current model and then fixes a target-specific batch. Cross-Benchmark Compression shares information across benchmarks and selects in advance. 
Adaptive Item Selection uses each new target result to choose the next item, while Sequential Decision and Stopping uses interim results to decide whether more testing is needed \citep{vivek2024anchor,yuan2025beyond,kipnis2025metabench,hofmann2025fluid,zhou2026celeus}. Using later and more target-specific information can improve selection, but reduces reuse and parallelism.

\subsection{Recovering the Full Result}

Random and stratified samples support recovery through means or group weights. A deterministic coreset score may instead describe a new short form. Inverse-probability weighting targets the original pool when sampling probabilities are known, though uneven weights increase variance. Historical predictors can recover untested results but depend on transfer to new models \citep{purificato2026select,polo2024tinybenchmarks,zhang2026sparseeval}. Residual or gold-label correction reduces dependence on prediction accuracy under suitable sampling conditions \citep{zhang2026benchmark,fisch2026collabeval}. Sequential methods may return only a winner, rank tier, or threshold decision. Comparisons must therefore match the recovered result.

\subsection{Lifecycle Cost}

Lifecycle cost includes subset construction, target-side signals, expensive tests, score recovery, updates, and latency. Content-Based methods avoid building a historical response matrix, although full-pool embeddings or reasoning traces still add cost. Historical-Response and Cross-Benchmark methods have higher construction and update costs, but low cost per new target when the same benchmark or suite is used repeatedly. Target-Conditioned methods pay for a probe, output scan, or cheap judge on every target, so the signal must be much cheaper than the tests or labels it removes \citep{polo2024tinybenchmarks,subramani2025simba,unell2025smarter}. Adaptive methods add serial rounds, while sequential stopping is cheap for clear cases but may approach full evaluation for close models. Item count, model calls, human review, judge calls, and wall-clock time should therefore be reported separately. Appendix~\ref{app:comparison-lifecycle} gives the detailed breakdown.

\subsection{Limitations and Method Choice}

Sections~\ref{sec:static} and~\ref{sec:adaptive} discuss the limitations of each family, with further analysis in Appendices~\ref{app:analysis} and~\ref{app:comparison}. These limitations guide method choice alongside the available information and required result. For a single benchmark, consider Content-Based methods when only item data are available, Historical-Response methods when past evaluation records are available, and Target-Conditioned methods when a cheap, useful target signal is available. Cross-Benchmark Compression suits related benchmarks evaluated jointly. Consider Adaptive Item Selection when serial rounds are acceptable and the score estimator accounts for how items are selected. Sequential Decision and Stopping requires a valid stopping rule and may save little for close models \citep{zhang2026benchmark,jiang2026can,arviv2026stop,zhou2026celeus}.
When a method's requirements are not met, use broader random sampling, test unsupported tasks directly, continue evaluation, or abstain. Appendix~\ref{app:comparison-choice} summarizes these checks and choices.

\section{Challenges and Future Directions}
\label{sec:challenges}

Beyond statistical reliability, benchmark compression must also generalize to future models and gain broader adoption in the evaluation community.

\paragraph{Generalization Across Models and Over Time.}
Many compressed subsets are selected using response records from existing models. Models from other families and newer, more capable models may show different response patterns, which may reduce the subset's coverage and discrimination. Future work should develop compression methods that transfer more reliably across models and over time. Compressed benchmarks should also be tested on newly released models and updated through periodic or continuous refresh mechanisms when their coverage or discrimination declines.

\paragraph{Updating Established Benchmarks.}
Although studies have found redundancy in established benchmarks, full results remain necessary for comparison with leaderboards and prior work. Researchers therefore often need to run complete evaluations for formal reporting. Compression still reduces costs during model iteration, screening, and repeated validation. A promising direction is to combine compression with benchmark repair and release a recalibrated version that removes flawed, redundant, overly easy, or weakly discriminating items. This could lower cost while improving discrimination among stronger models, providing a cheaper and more informative shared reference.

\paragraph{Compression Before Release.}
For new benchmarks, compression should be integrated into benchmark construction before release. The public version should use a compact but informative item set that covers the intended model capabilities from the outset, rather than publishing a large redundant benchmark and compressing it later. This compression by design can accelerate evaluation and model iteration while reducing unnecessary computational and environmental costs.

\section{Conclusion}
\label{sec:conclusion}

In this survey, we introduced a unified taxonomy for benchmark compression in LLM evaluation. We also compared method families by their information needs, costs, and reliability. Our analysis shows that methods reduce costs by evaluating representative items, using cheaper or reusable information to estimate omitted results, and focusing further evaluation on uncertain cases. Reliable compression requires sufficient coverage, accurate recovery, and valid adaptive decisions. More broadly, efficient evaluation should collect useful information at lower cost while preserving reliable results.

\section*{Limitations}

This survey has three limitations. First, it covers work only through August 2026, so subsequent studies are not included. We will continue to update the survey as new research appears. Second, space constraints and the rapid growth of benchmark compression prevent the main text from describing every method in detail. We therefore focus on a high-level taxonomy and roadmap that help readers quickly understand the field as a whole, its main method families, and their relationships. Appendix~\ref{app:analysis} provides detailed descriptions of representative methods, while Appendix~\ref{app:comparison} presents in-depth comparisons across method families. Third, studies use widely different evaluation settings, making their results difficult to compare or combine quantitatively. Fortunately, our common formulation and comparison framework make these differences explicit, supporting more consistent reporting and future quantitative synthesis.

\bibliography{custom}

\clearpage
\appendix
\definecolor{taxsingle}{HTML}{4472C4}
\definecolor{taxcross}{HTML}{70AD47}
\definecolor{taxadaptive}{HTML}{E8A2B8}
\definecolor{taxfoundation}{HTML}{7B91AE}
\colorlet{appheader}{taxsingle!12}
\colorlet{appstripe}{taxsingle!3}
\renewcommand{\yes}{\textcolor{taxsingle!85!black}{$\checkmark$}}

\flushbottom
\setcounter{dbltopnumber}{3}
\renewcommand{\dbltopfraction}{0.95}
\renewcommand{\textfraction}{0.05}

\begingroup
\etocdepthtag{appendix}
\etocsettagdepth{main}{none}
\etocsettagdepth{appendix}{subsubsection}
\etocsettocstyle{\section*{Appendix Contents}}{}
\newcommand{\tocdot}{\nobreak\leaders\hbox to 0.8em{\hss.\hss}\hfill}
\etocsetstyle{section}
  {\par}
  {\addvspace{0.45em}}
  {\noindent \textbf{\etocnumber{} \etocname} \tocdot \etocpage \par}
  {}
\etocsetstyle{subsection}
  {}
  {\addvspace{0.20em}}
  {\noindent \hspace{0.5em} \etocnumber{} \etocname \tocdot \etocpage \par}
  {}
\etocsetstyle{subsubsection}
  {}
  {\addvspace{0.15em}}
  {\noindent \hspace{1.5em} \etocnumber{} \etocname \tocdot \etocpage \par}
  {}
\normalsize
\setlength{\parskip}{3.5pt}
\setlength{\parindent}{0pt}
\tableofcontents
\endgroup
\newpage

\section{Paper Catalogue}
\label{app:catalogue}

The catalogue maps individual papers to the taxonomy. The complete tables support paper lookup and corpus counts, while the boundary examples illustrate how the classification rules handle methods with multiple properties. Appendices~\ref{app:analysis} and~\ref{app:comparison} define the categories and compare their information requirements, costs, and recovery rules.

\subsection{Reading the Catalogue}
\label{app:catalogue-scope}

Each paper appears once under the primary label defined in Appendix~\ref{app:taxonomy}, enabling easy lookup and stable category counts. The label captures the paper's main contribution rather than every mechanism it uses. Table~\ref{tab:taxonomy-attributes} records the secondary attributes of plan timing, benchmark scope, main information, saved unit, and returned result. The four catalogue tables further report Domain, Venue, Year, GitHub, Saved Unit, and Conclusion. We use the authors' method name when available and otherwise derive a short label from the title. Venue and year follow the verified formal publication, or a verified submission or preprint when no formal version exists.

Domain codes are Gen for general, Rsn for reasoning, MM for multimodal, MT for translation, IR for retrieval, SE for software engineering, Know. for knowledge, and Sec for security. ACLF and EMNLPF denote Findings of ACL and Findings of EMNLP. Other venue and task codes are standard or defined in the captions. The saved unit identifies the costly operation reduced, while S, R, D, and CI denote score, ranking, decision, and reliability interval. A GitHub icon marks a verified repository, and N/A means that none was found. Short-form rows should be read narrowly. An operational label covers only the score, ranking, or decision evaluated in the paper. A filtered-subset score does not estimate the full-benchmark mean without a recovery rule. Work that defines a new measurement target or lacks a fixed full reference belongs to Benchmark Analysis and Foundations or falls outside our scope. Source-to-target methods must also pass a cost check, as replacing target tests lowers marginal cost only when source results are already available or cheaper to obtain.

\begingroup
\makeatletter
\global\setlength{\@dblfptop}{0pt}
\global\setlength{\@dblfpsep}{7pt}
\global\setlength{\@dblfpbot}{0pt plus 1fil}
\makeatother

\begin{table*}[p]
\centering
\fontsize{8}{10}\selectfont
\setlength{\tabcolsep}{1.20pt}
\renewcommand{\arraystretch}{1.05}
\begin{tabular}{@{}L{4.80cm}L{3.00cm}L{1.60cm}>{\centering\arraybackslash}p{0.90cm}>{\centering\arraybackslash}p{1.05cm}>{\centering\arraybackslash}p{1.65cm}>{\centering\arraybackslash}p{1.70cm}@{}}
\toprule
\rowcolor{taxsingle!12}Method label & Domain & Venue & Year & GitHub & Saved unit & Conclusion \\
\midrule
\rowcolor{taxsingle!7}\multicolumn{7}{@{}l}{\textcolor{taxsingle!85!black}{\textbf{Content-Based Compression (10)}}} \\
\addlinespace[1pt]
Scales++~\citep{bean2025scales++} & Gen/QA/Rsn/Math & arXiv & 2025 & N/A & Item & S \\
PromptEval~\citep{polo2024efficient} & Gen/QA/Rsn & NeurIPS & 2024 & \href{\detokenize{https://github.com/felipemaiapolo/prompteval}}{\textcolor{black!80}{\faGithub}} & Prompt & S \\
EffiEval~\citep{wang2025effieval} & Gen/QA/Rsn/Math & arXiv & 2025 & N/A & Item & S/R \\
MIS Select~\citep{kjorvezir2026consistent} & Gen/Rsn/Math & arXiv & 2026 & N/A & Prompt & R/CI \\
Agreement~\citep{purificato2026select} & Gen/NLG & arXiv & 2026 & N/A & Human & S \\
PRECISE~\citep{divekar2026statistically} & IR & GEM@ACL & 2026 & N/A & Human & S/R/CI \\
CMLE~\citep{shen2026robust} & Safety & arXiv & 2026 & N/A & Human & S \\
NoisyHT~\citep{feng2026noisy} & Safety & ICLR & 2026 & N/A & Human & D \\
ARES~\citep{saad2024ares} & RAG/IR & NAACL & 2024 & \href{\detokenize{https://github.com/stanford-futuredata/ARES}}{\textcolor{black!80}{\faGithub}} & Human & S/R/CI \\
CoT-Core~\citep{pan2026cot} & Gen/QA/Rsn/Math & arXiv & 2026 & N/A & Item & S/R \\
\midrule
\rowcolor{taxsingle!7}\multicolumn{7}{@{}l}{\textcolor{taxsingle!85!black}{\textbf{Historical-Response Compression (26)}}} \\
\addlinespace[1pt]
AnchorPoints~\citep{vivek2024anchor} & Gen/QA/Rsn & EACL & 2024 & N/A & Item & S/R \\
tinyBenchmarks~\citep{polo2024tinybenchmarks} & Gen/QA/Rsn/Math & ICML & 2024 & \href{\detokenize{https://github.com/felipemaiapolo/tinyBenchmarks}}{\textcolor{black!80}{\faGithub}} & Item+judge & S/R \\
SubLIME~\citep{saranathan2025sublime} & Gen/QA/Rsn/Math & ACL & 2025 & N/A & Item & S/R \\
MiniLongBench~\citep{huang2025minilongbench} & Gen/QA/IR & ACL & 2025 & N/A & Item & S/R \\
DataLite~\citep{wang2026rethinking} & Gen/QA/Rsn/Math & ICLR & 2026 & N/A & Item & S/R \\
SparseEval~\citep{zhang2026sparseeval} & Gen/QA/Rsn/Math & ICLR & 2026 & \href{\detokenize{https://github.com/taolinzhang/SparseEval}}{\textcolor{black!80}{\faGithub}} & Item & S/R \\
REP-CORE~\citep{zhang2026learning} & Gen/QA/Rsn/MM & ICML & 2026 & N/A & Item & S/R \\
SMART~\citep{gupta2025improving} & Gen/QA/Rsn & NAACL & 2025 & \href{\detokenize{https://github.com/facebookresearch/ResponsibleNLP/tree/main/SMART-Filtering}}{\textcolor{black!80}{\faGithub}} & Item & S/R \\
AIPW~\citep{zhang2026benchmark} & Gen/QA/Rsn & NeurIPS & 2025 & \href{\detokenize{https://github.com/socialfoundations/benchmark-prediction}}{\textcolor{black!80}{\faGithub}} & Item & S/R \\
MINCE~\citep{das2026mince} & Gen/QA/Rsn/Math & arXiv & 2026 & N/A & Item & S/R \\
DISCO~\citep{rubinstein2026disco} & Gen/QA/Rsn & ICLR & 2026 & N/A & Item & S/R \\
mRMR++~\citep{bowyer2026efficient} & Gen/QA/Rsn/Code & arXiv & 2026 & \href{\detokenize{https://github.com/sambowyer/mrmr_eval}}{\textcolor{black!80}{\faGithub}} & Query & S/R \\
Option IRT~\citep{fei2026every} & Gen/QA/Rsn/Math & arXiv & 2026 & N/A & MCQ & S/R \\
CollabEval~\citep{fisch2026collabeval} & Gen/QA/Code/Human & arXiv & 2026 & N/A & Prompt & S/CI \\
BiasBound~\citep{zhuang2025efficient} & Gen/QA/Rsn & TPAMI & 2025 & N/A & Item & S/R \\
ShortForm~\citep{gignac2025psychometrically} & Gen/QA/Rsn/Math & Intelligence & 2025 & N/A & Human & S/R \\
LURE~\citep{berrada2026scaling} & Gen/QA/Rsn & NeurIPS & 2025 & \href{\detokenize{https://github.com/gabrielleberrada/scaling-up-active-testing}}{\textcolor{black!80}{\faGithub}} & Item+label & S/CI \\
MetaEval~\citep{wang2026metaeval} & Gen/QA/Rsn/Math & AAAI & 2026 & \href{\detokenize{https://github.com/wangzhuo0092/MetaEval}}{\textcolor{black!80}{\faGithub}} & Item & S/R \\
Neyman~\citep{liu2026active} & Gen/QA/Rsn/MM & arXiv & 2026 & N/A & Item & S \\
DatBench~\citep{joshi2026datbench} & MM/VQA & arXiv & 2026 & \href{\detokenize{https://github.com/datologyai/DatBench}}{\textcolor{black!80}{\faGithub}} & VLM item & S/R \\
SoftElo~\citep{kargi2026uncertain} & Chat/Human & arXiv & 2026 & \href{\detokenize{https://github.com/kargibora/SoftElo}}{\textcolor{black!80}{\faGithub}} & Pair & S/R/CI \\
MidRange~\citep{ndzomga2026efficient} & Code/Agent & arXiv & 2026 & \href{\detokenize{https://github.com/fsndzomga/efficient-benchmarking-ai-agents}}{\textcolor{black!80}{\faGithub}} & Agent & S/R \\
CJE~\citep{landesberg2025causal} & Gen/Chat/Code/Human & arXiv & 2025 & \href{\detokenize{https://github.com/cimo-labs/cje}}{\textcolor{black!80}{\faGithub}} & Human & S/R/CI \\
BISS~\citep{matricon2025efficiently} & Code/SE & arXiv & 2025 & N/A & Test & R \\
JBDistill~\citep{zhang2025jailbreak} & Safety/Sec & EMNLPF & 2025 & N/A & Jailbreak & S/R/CI \\
Optimal Sampling~\citep{li2025towards} & Gen/QA/Rsn/Math & EMNLP & 2025 & N/A & Query & S \\
\midrule
\rowcolor{taxsingle!7}\multicolumn{7}{@{}l}{\textcolor{taxsingle!85!black}{\textbf{Target-Conditioned Compression (15)}}} \\
\addlinespace[1pt]
TAILOREDBENCH~\citep{yuan2025beyond} & Gen/Math/MM & ACL & 2025 & \href{\detokenize{https://github.com/marvelcell/TailoredBench}}{\textcolor{black!80}{\faGithub}} & Item & S/R \\
GAT~\citep{ramakrishnan2026generative} & Gen/QA/Med & arXiv & 2026 & N/A & Human & S \\
MAD~\citep{feng2025sample} & Gen/Math/Code/Human & ACL & 2025 & \href{\detokenize{https://github.com/weiji-Feng/MAD-Eval}}{\textcolor{black!80}{\faGithub}} & Pair & R \\
MetricMatch~\citep{unell2026metric} & Gen/Human & arXiv & 2026 & \href{\detokenize{https://github.com/som-shahlab/MetricMatch}}{\textcolor{black!80}{\faGithub}} & Human & D/CI \\
Judge Sampling~\citep{unell2025smarter} & Gen/Human & LLM-Eval & 2025 & N/A & Human & S/CI \\
ReviewAug~\citep{kim2026augmenting} & Med/Human & arXiv & 2026 & N/A & Human & S/CI \\
JudgeAdjust~\citep{lee2025correctly} & Chat/Human & ICML & 2026 & \href{\detokenize{https://github.com/UW-Madison-Lee-Lab/LLM-judge-reporting}}{\textcolor{black!80}{\faGithub}} & Human & S/R/CI \\
RidgePPI~\citep{eyre2024regression} & Gen & ICML & 2025 & N/A & Human & S \\
QuEst~\citep{deng2025quest} & Gen/Safety/NLG & ICML & 2025 & \href{\detokenize{https://github.com/btleyre/quest}}{\textcolor{black!80}{\faGithub}} & Judge & S/R/D/CI \\
GLIDE~\citep{martinon2026industrializing} & Agent/Safety & arXiv & 2026 & \href{\detokenize{https://github.com/EmertonData/glide}}{\textcolor{black!80}{\faGithub}} & Human & S/CI \\
CostEval~\citep{angelopoulos2025cost} & Gen/Chat & arXiv & 2025 & N/A & Human & S \\
ControlEval~\citep{zhou2025accelerating} & Chat/Human & ICML & 2025 & \href{\detokenize{https://github.com/Zanette-Labs/control_variates_evaluation}}{\textcolor{black!80}{\faGithub}} & Human & S \\
AutoEval~\citep{boyeau2024autoeval} & Gen/Chat & ICML & 2025 & \href{\detokenize{https://github.com/PierreBoyeau/autoeval}}{\textcolor{black!80}{\faGithub}} & Human & S/R/CI \\
SSME~\citep{shanmugam2026evaluating} & Gen/NLP & NeurIPS & 2025 & \href{\detokenize{https://github.com/divyashan/SSME}}{\textcolor{black!80}{\faGithub}} & Label & S \\
DiffUse~\citep{tahan2024label} & Gen/NLG/Human & ACL & 2024 & N/A & Pair & D \\
\bottomrule
\end{tabular}
\caption{\textbf{Catalogue of single-benchmark compression methods.} S, R, D, and CI denote score, ranking, decision, and reliability interval. EMNLPF denotes Findings of EMNLP.}
\label{tab:catalogue}
\end{table*}

\begin{table*}[p]
\centering
\fontsize{8}{10}\selectfont
\setlength{\tabcolsep}{1.2pt}
\renewcommand{\arraystretch}{1.05}
\begin{tabular}{@{}L{4.80cm}L{3.00cm}L{1.60cm}>{\centering\arraybackslash}p{0.90cm}>{\centering\arraybackslash}p{1.05cm}>{\centering\arraybackslash}p{1.65cm}>{\centering\arraybackslash}p{1.70cm}@{}}
\toprule
\rowcolor{taxcross!12}Method label & Domain & Venue & Year & GitHub & Saved unit & Conclusion \\
\midrule
metabench~\citep{kipnis2025metabench} & Gen/QA/Rsn/Math & ICLR & 2025 & \href{\detokenize{https://github.com/adkipnis/metabench}}{\textcolor{black!80}{\faGithub}} & Item & S/R/CI \\
Adaptive Sampling~\citep{xu2024data} & Gen/Code & arXiv & 2024 & N/A & Item & S/R \\
LEGO-IRT~\citep{liao2025toward} & Gen/QA/Know./MT & arXiv & 2025 & \href{\detokenize{https://github.com/Rorschach1989/efficient-lm-eval}}{\textcolor{black!80}{\faGithub}} & Item & S/R/CI \\
FPC~\citep{habba2026growing} & Gen/QA/Rsn/Math & arXiv & 2026 & \href{\detokenize{https://github.com/eliyahabba/growing-pains}}{\textcolor{black!80}{\faGithub}} & Item & S/R \\
Facility~\citep{yao2026coresets} & Gen & arXiv & 2026 & N/A & Item & S/R \\
Agent PACE~\citep{song2026pace} & Rsn/Code/Agent & arXiv & 2026 & N/A & Agent & S/R \\
Submodular~\citep{smola2026submodular} & Gen & arXiv & 2026 & \href{\detokenize{https://github.com/smolix/benchmark-selection}}{\textcolor{black!80}{\faGithub}} & Run & S/CI \\
SimBA~\citep{subramani2025simba} & Gen/QA/Rsn/Math & EMNLPF & 2025 & \href{\detokenize{https://github.com/nishantsubramani/simba}}{\textcolor{black!80}{\faGithub}} & Run & S/R \\
SkillFactors~\citep{maimon2026benchmarks} & Gen & TACL & 2026 & N/A & Run & S/R/D \\
BIS~\citep{shi2025importance} & Code & arXiv & 2025 & N/A & Run & S \\
PSN-IRT~\citep{zhou2026lost} & Gen/QA/Rsn/Math & AAAI & 2026 & \href{\detokenize{https://github.com/Joe-Hall-Lee/PSN-IRT}}{\textcolor{black!80}{\faGithub}} & Item & R \\
MultiIRT~\citep{lior2026extending} & Gen/QA/Know./MT & arXiv & 2026 & N/A & Item-lang. & S/CI \\
BenchPress~\cite{zeng2026you} & Gen & arXiv & 2026 & \href{\detokenize{https://github.com/microsoft/benchpress2}}{\textcolor{black!80}{\faGithub}} & Run & S/R/CI \\
ResampledBench~\cite{suzuki2025resampling} & MM/VQA & arXiv & 2025 & N/A & Item & R \\
\bottomrule
\end{tabular}
\caption{\textbf{Catalogue of cross-benchmark compression methods.} Conclusion abbreviations follow Table~\ref{tab:catalogue}. EMNLPF denotes Findings of EMNLP.}
\label{tab:catalogue-cross}
\end{table*}

\begin{table*}[p]
\centering
\fontsize{8}{10}\selectfont
\setlength{\tabcolsep}{1.2pt}
\renewcommand{\arraystretch}{1.05}
\begin{tabular}{@{}L{4.80cm}L{3.00cm}L{1.60cm}>{\centering\arraybackslash}p{0.90cm}>{\centering\arraybackslash}p{1.05cm}>{\centering\arraybackslash}p{1.65cm}>{\centering\arraybackslash}p{1.70cm}@{}}
\toprule
\rowcolor{taxadaptive!15}Method label & Domain & Venue & Year & GitHub & Saved unit & Conclusion \\
\midrule
\rowcolor{taxadaptive!7}\multicolumn{7}{@{}l}{\textcolor{taxadaptive!65!black}{\textbf{Adaptive Item Selection (21)}}} \\
\addlinespace[1pt]
AEA~\citep{li2024active} & Gen/QA/Rsn & ICML & 2025 & N/A & Item+judge & S \\
Fluid~\citep{hofmann2025fluid} & Gen/QA/Rsn/Math & COLM & 2025 & \href{\detokenize{https://github.com/allenai/fluid-benchmarking}}{\textcolor{black!80}{\faGithub}} & Item & S/R \\
ATLAS~\citep{li2025adaptive} & Gen/QA/Rsn/Math & ICML & 2026 & \href{\detokenize{https://github.com/Peiyu-Georgia-Li/ATLAS}}{\textcolor{black!80}{\faGithub}} & Item & S/R/CI \\
\papercitealias{FactorCAT}{truongbeyond}{2025} & Gen/QA/Rsn & ICLR Sub. & 2025 & N/A & Item & S/R/CI \\
V-Optimal~\citep{krumdick2026cost} & Gen/QA/Rsn/Math & arXiv & 2026 & N/A & Item & S \\
ReEval~\citep{truong2025reliable} & Gen/QA/Rsn & ICML & 2025 & \href{\detokenize{https://github.com/sangttruong/reeval}}{\textcolor{black!80}{\faGithub}} & Item & S/CI \\
M3IRT~\citep{uebayashi2026evaluating} & Rsn/MM/VQA & ICLR & 2026 & \href{\detokenize{https://github.com/CyberAgentAILab/M3IRT}}{\textcolor{black!80}{\faGithub}} & Item & S/R \\
BayesAME~\citep{cordero2026bayesame} & Gen/QA/Rsn/Math & arXiv & 2026 & N/A & Item & S/R/CI \\
IRSL~\citep{truong2026item} & Gen/QA/Rsn/Math & ICML & 2026 & \href{\detokenize{https://github.com/aims-foundations/irsl}}{\textcolor{black!80}{\faGithub}} & Item & S/R/D \\
AcTracer~\citep{huang2026actracer} & Gen/QA/Rsn/Math & TOSEM & 2026 & \href{\detokenize{https://github.com/YuhengHuang42/test_selection_llm}}{\textcolor{black!80}{\faGithub}} & Judge & S \\
AutoJudger~\citep{ding2026autojudger} & MM/VQA & ACL & 2026 & \href{\detokenize{https://github.com/IMNearth/AutoJudger}}{\textcolor{black!80}{\faGithub}} & Item & S/R \\
Cer-Eval~\citep{wang2025cer} & Gen/QA/Rsn/Math & arXiv & 2025 & N/A & Item & S/D/CI \\
FAQ-PAI~\citep{wu2026efficient} & Gen/QA/Rsn/Math & arXiv & 2026 & \href{\detokenize{https://github.com/skbwu/efficiently-evaluating-llms}}{\textcolor{black!80}{\faGithub}} & Item & S/CI \\
CEM~\citep{kim2026measuring} & Math & arXiv & 2026 & N/A & Item & CI \\
RIPr~\citep{hsu2026efficient} & Gen/QA/Rsn/Math & arXiv & 2026 & N/A & Item & S/CI \\
IRT-Agent~\citep{ge2026agent} & Code/Agent & arXiv & 2026 & \href{\detokenize{https://github.com/dariakryvosheieva/agent-psychometrics}}{\textcolor{black!80}{\faGithub}} & Run & S \\
APOT~\citep{asgari2025adaptive} & Gen & AST & 2025 & N/A & Item & S/CI \\
SafetyCAT~\citep{spagliardi2026efficient} & Safety & WS@ICML & 2026 & N/A & Item & S/R \\
CollabFilter~\citep{zhong2025efficient} & Gen/QA/Rsn/Math & arXiv & 2025 & N/A & Item & S/R \\
ROBIN-HOOD~\citep{saha2026llm} & Human & arXiv & 2026 & N/A & Judge & S \\
FAIR-Eval~\citep{anonymous2026faireval} & Gen/QA/Rsn/Math & ARR Sub. & 2026 & N/A & Item & S/R/CI \\
\midrule
\rowcolor{taxadaptive!7}\multicolumn{7}{@{}l}{\textcolor{taxadaptive!65!black}{\textbf{Sequential Decision and Stopping (15)}}} \\
\addlinespace[1pt]
Flash-HELM~\citep{perlitz2024efficient} & Gen/QA/Rsn & NAACL & 2024 & N/A & Prompt & S/R/D/CI \\
UCB-E~\citep{zhou2025speeding} & Gen/QA/Rsn/Math & ICLR & 2025 & \href{\detokenize{https://github.com/kilian-group/banditeval}}{\textcolor{black!80}{\faGithub}} & Example & R/D \\
AdaptiveRank~\citep{balkir2026confident} & Gen/QA/MT/NLG & arXiv & 2026 & \href{\detokenize{https://github.com/trismik/continuous-cat}}{\textcolor{black!80}{\faGithub}} & Item & S/R/CI \\
JustEnough~\citep{arviv2026stop} & MM/Vision & ACLF & 2026 & \href{\detokenize{https://github.com/OfirArviv/adaptive-eval}}{\textcolor{black!80}{\faGithub}} & Item & S/R/D/CI \\
SySRs~\citep{lyu2026cutting} & Gen/QA/Rsn/Math & ICML & 2026 & \href{\detokenize{https://github.com/zifanlyu/llm-bandits-sysrs}}{\textcolor{black!80}{\faGithub}} & Query & D \\
SELECT-LLM~\citep{durmazkeser2026large} & Gen/QA/Rsn/Math & arXiv & 2026 & N/A & Human & D \\
CELEUS~\citep{zhou2026celeus} & Gen/QA & arXiv & 2026 & \href{\detokenize{https://github.com/zyecs/celeus}}{\textcolor{black!80}{\faGithub}} & Item & S/CI \\
ParEvalLayer~\citep{huang2026parevallayer} & Agent/SE & arXiv & 2026 & N/A & Agent & D/CI \\
PULSE~\citep{tolochinsky2026valid} & Gen/QA/Rsn/Math & arXiv & 2026 & \href{\detokenize{https://github.com/elad-tolo/pulse-llm-eval}}{\textcolor{black!80}{\faGithub}} & Example & D \\
TrustEscalate~\citep{jung2025trust} & Chat/NLG/Human & ICLR & 2025 & N/A & Judge & CI \\
AV-AIVAT~\citep{li2026av} & Agent & arXiv & 2026 & N/A & Game hand & S/R/CI \\
R-AutoEval+~\citep{park2026adaptive} & Gen/QA/Rsn/Math & NeurIPS & 2025 & \href{\detokenize{https://github.com/kclip/R_AutoEval_plus}}{\textcolor{black!80}{\faGithub}} & Human & D/CI \\
HiddenGems~\citep{kahana2026discovering} & Gen/QA/Rsn/Math & arXiv & 2026 & N/A & Query & D \\
Arena-Lite~\citep{son2025arena} & Chat/Human & EMNLP & 2025 & N/A & Judge & S/R \\
De-Arena~\citep{yin2026decentralized} & Gen/Chat/Rsn/Math & ACL & 2026 & N/A & Judge & S/R \\
\bottomrule
\end{tabular}
\caption{\textbf{Catalogue of adaptive evaluation methods.} Conclusion abbreviations follow Table~\ref{tab:catalogue}. ACLF denotes Findings of ACL.}
\label{tab:catalogue-adaptive}
\end{table*}

\begin{table*}[t]
\centering
\fontsize{8}{10}\selectfont
\setlength{\tabcolsep}{1.6pt}
\renewcommand{\arraystretch}{1.05}
\begin{tabular}{@{}L{4.80cm}L{3.00cm}L{1.60cm}>{\centering\arraybackslash}p{0.90cm}>{\centering\arraybackslash}p{1.05cm}L{3.40cm}@{}}
\toprule
\rowcolor{taxfoundation!12}Method label & Domain & Venue & Year & GitHub & Analysis focus \\
\midrule
ONEBench~\citep{ghosh2025onebench} & Gen/MM & ACL & 2025 & N/A & Sparse ranking \\
DualEval~\citep{li2026dualeval} & Gen/Math/Code/Know. & arXiv & 2026 & \href{\detokenize{https://github.com/aaron-jx-li/DualEval}}{\textcolor{black!80}{\faGithub}} & Item calibration \\
Redundancy~\citep{zhang2025redundancy} & MM/Vision & ACL & 2025 & N/A & Benchmark redundancy \\
IRT Audit~\citep{jiang2026can} & Gen/Rsn/Math/Know. & arXiv & 2026 & N/A & IRT reliability \\
TaskBudget~\citep{huang2026many} & Code/Agent & arXiv & 2026 & \href{\detokenize{https://github.com/WilliamWJHuang/How-Many-Tasks-Are-Enough-for-Agent-Benchmark-Decisions}}{\textcolor{black!80}{\faGithub}} & Task sufficiency \\
cBMM~\citep{qu2026interpretable} & Gen/Rsn/Math/Know. & arXiv & 2026 & N/A & Latent calibration \\
JudgeCal~\citep{li2026calibrate} & Gen/Human & arXiv & 2026 & N/A & Judge calibration \\
JudgeDiag~\citep{gupta2026diagnosing} & Gen/NLG/Human & arXiv & 2026 & N/A & Judge reliability \\
HarnessAudit~\citep{zhang2026stop} & Code/Agent & arXiv & 2026 & \href{\detokenize{https://github.com/yunbeizhang/harness-binding-constraint}}{\textcolor{black!80}{\faGithub}} & Harness sensitivity \\
BestN~\citep{landesberg2026llm} & Chat/Human & arXiv & 2026 & N/A & Best-of-N validity \\
BenchScope~\citep{sha2026benchscope} & Gen/Code/Agent & arXiv & 2026 & N/A & Effective dimension \\
OPD~\citep{cao2026does} & Gen/QA/Rsn/Know. & AAAI & 2026 & N/A & Answer bias \\
LaplaceIRT~\citep{francisco2026laplace} & Gen/QA/Rsn/Math & arXiv & 2026 & \href{\detokenize{https://github.com/JFMandujanoR/Laplace-PSN-IRT/tree/jfmr_laplace}}{\textcolor{black!80}{\faGithub}} & IRT uncertainty \\
MDAD~\citep{yauney2026reliable} & Gen/QA/Rsn/Math & ICLR & 2026 & \href{\detokenize{https://github.com/dill-lab/micro-benchmarking-reliability}}{\textcolor{black!80}{\faGithub}} & Micro-benchmark reliability \\
ReliableEval~\citep{lior2025reliableeval} & Gen/QA/Rsn/Math & EMNLPF & 2025 & \href{\detokenize{https://github.com/SLAB-NLP/Reliable-Eval}}{\textcolor{black!80}{\faGithub}} & Prompt variability \\
Resolution~\citep{kotawala2026resolution} & Gen/QA/Rsn/Math & arXiv & 2026 & \href{\detokenize{https://github.com/akotawala10/llm-power}}{\textcolor{black!80}{\faGithub}} & Ranking resolution \\
ErrorBars~\citep{miller2024adding} & Gen/QA/Math/Code & arXiv & 2024 & N/A & Evaluation uncertainty \\
BENCHMARK2~\citep{qian2026benchmark} & Gen/Rsn/Math/Know. & arXiv & 2026 & N/A & Benchmark quality \\
\bottomrule
\end{tabular}
\caption{\textbf{Catalogue of benchmark analysis and foundations papers.} EMNLPF denotes Findings of EMNLP.}
\label{tab:catalogue-foundations}
\end{table*}

\endgroup
\FloatBarrier

\subsection{Boundary Examples}
\label{app:catalogue-reading}

Table~\ref{tab:catalogue-boundaries} applies the ordered rules to papers that combine several properties. The deciding reason follows the primary contribution, while the last column keeps the other properties visible. These examples clarify the labels without changing the catalogue counts.

\begin{table*}[t]
\centering
\footnotesize
\setlength{\tabcolsep}{5pt}
\renewcommand{\arraystretch}{1.18}
\arrayrulecolor{black!22}
\begin{tabularx}{\textwidth}{@{}>{\raggedright\arraybackslash}p{2.05cm}>{\raggedright\arraybackslash}p{4.05cm}X@{}}
\toprule
\rowcolor{appheader}
Paper & Category boundary & Classification basis \\
\midrule
\rowcolor{appstripe}
TAILORED\newline BENCH\newline\citep{yuan2025beyond} &
{\scriptsize\textcolor{black!58}{\textbf{Possible overlap}}}\par\vspace{0.5pt}
{\fontsize{7}{8}\selectfont\textcolor{black}{Target-Conditioned vs. Adaptive}}\par\smallskip
{\scriptsize\textcolor{black!58}{\textbf{Primary label}}}\par\vspace{0.5pt}
{\fontsize{8}{9}\selectfont Target-Conditioned Compression} &
\textcolor{black!58}{\textbf{Deciding reason}}\enspace A small target probe chooses the source neighbors and one target-specific batch. Results from that main batch do not revise later tests or trigger stopping.\par\smallskip
\textcolor{black!58}{\textbf{Secondary attributes}}\enspace one target-conditioned batch, single benchmark, historical results and current model signal, item, score and ranking \\
\addlinespace[2pt]
metabench~\citep{kipnis2025metabench} &
{\scriptsize\textcolor{black!58}{\textbf{Possible overlap}}}\par\vspace{0.5pt}
{\fontsize{7}{8}\selectfont\textcolor{black}{Cross-Benchmark vs. Historical-Response}}\par\smallskip
{\scriptsize\textcolor{black!58}{\textbf{Primary label}}}\par\vspace{0.5pt}
{\fontsize{8}{9}\selectfont Cross-Benchmark Compression} &
\textcolor{black!58}{\textbf{Deciding reason}}\enspace Historical results guide selection and recovery, but one sparse design jointly covers a benchmark suite and recovers both constituent and suite results.\par\smallskip
\textcolor{black!58}{\textbf{Secondary attributes}}\enspace fixed, joint suite, historical results, item, score, ranking, and interval \\
\addlinespace[2pt]
\rowcolor{appstripe}
PACE~\citep{song2026pace} &
{\scriptsize\textcolor{black!58}{\textbf{Possible overlap}}}\par\vspace{0.5pt}
{\fontsize{7}{8}\selectfont\textcolor{black}{Cross-Benchmark vs. Historical-Response}}\par\smallskip
{\scriptsize\textcolor{black!58}{\textbf{Primary label}}}\par\vspace{0.5pt}
{\fontsize{8}{9}\selectfont Cross-Benchmark Compression} &
\textcolor{black!58}{\textbf{Deciding reason}}\enspace Historical model results train the predictor, but the evaluation plan replaces expensive target agent runs with a fixed set of cheaper source items.\par\smallskip
\textcolor{black!58}{\textbf{Secondary attributes}}\enspace fixed, source to target, historical results, agent run, score and ranking \\
\addlinespace[2pt]
BIS~\citep{shi2025importance} &
{\scriptsize\textcolor{black!58}{\textbf{Possible overlap}}}\par\vspace{0.5pt}
{\fontsize{7}{8}\selectfont\textcolor{black}{Cross-Benchmark vs. Foundations}}\par\smallskip
{\scriptsize\textcolor{black!58}{\textbf{Primary label}}}\par\vspace{0.5pt}
{\fontsize{8}{9}\selectfont Cross-Benchmark Compression} &
\textcolor{black!58}{\textbf{Deciding reason}}\enspace Source-benchmark outcomes estimate a target-benchmark score. This replaces target tests, but it reduces marginal model inference only when the source is cached or cheaper.\par\smallskip
\textcolor{black!58}{\textbf{Secondary attributes}}\enspace fixed, source to target, prompt features, target test, score \\
\addlinespace[2pt]
\rowcolor{appstripe}
Fluid~\citep{hofmann2025fluid} &
{\scriptsize\textcolor{black!58}{\textbf{Possible overlap}}}\par\vspace{0.5pt}
{\fontsize{7}{8}\selectfont\textcolor{black}{Adaptive Item Selection vs. Sequential Decision and Stopping}}\par\smallskip
{\scriptsize\textcolor{black!58}{\textbf{Primary label}}}\par\vspace{0.5pt}
{\fontsize{8}{9}\selectfont Adaptive Item Selection} &
\textcolor{black!58}{\textbf{Deciding reason}}\enspace Each target response updates ability and determines the next item. The main contribution is item selection and latent-score recovery, so stopping remains secondary.\par\smallskip
\textcolor{black!58}{\textbf{Secondary attributes}}\enspace adaptive, single benchmark, historical and online results, item, score and ranking \\
\addlinespace[2pt]
CELEUS~\citep{zhou2026celeus} &
{\scriptsize\textcolor{black!58}{\textbf{Possible overlap}}}\par\vspace{0.5pt}
{\fontsize{7}{8}\selectfont\textcolor{black}{Adaptive Item Selection vs. Sequential Decision and Stopping}}\par\smallskip
{\scriptsize\textcolor{black!58}{\textbf{Primary label}}}\par\vspace{0.5pt}
{\fontsize{8}{9}\selectfont Sequential Decision and Stopping} &
\textcolor{black!58}{\textbf{Deciding reason}}\enspace The main guarantee is an anytime-valid interval and a valid stop when the target precision is reached.\par\smallskip
\textcolor{black!58}{\textbf{Secondary attributes}}\enspace adaptive, single benchmark, current model and online results, item, score and interval \\
\addlinespace[2pt]
\rowcolor{appstripe}
SMART~\citep{gupta2025improving} &
{\scriptsize\textcolor{black!58}{\textbf{Possible overlap}}}\par\vspace{0.5pt}
{\fontsize{7}{8}\selectfont\textcolor{black}{Compression vs. Benchmark Repair}}\par\smallskip
{\scriptsize\textcolor{black!58}{\textbf{Primary label}}}\par\vspace{0.5pt}
{\fontsize{8}{9}\selectfont Historical-Response Compression} &
\textcolor{black!58}{\textbf{Deciding reason}}\enspace It starts from fixed benchmarks and checks whether a reusable filtered subset preserves the original model order. Its short-set score is not a recovered full-benchmark mean.\par\smallskip
\textcolor{black!58}{\textbf{Secondary attributes}}\enspace fixed, single benchmark, item content and historical results, item, score and ranking \\
\addlinespace[2pt]
ONEBench\newline\citep{ghosh2025onebench} &
{\scriptsize\textcolor{black!58}{\textbf{Possible overlap}}}\par\vspace{0.5pt}
{\fontsize{7}{8}\selectfont\textcolor{black}{Operational Method vs. Foundations}}\par\smallskip
{\scriptsize\textcolor{black!58}{\textbf{Primary label}}}\par\vspace{0.5pt}
{\fontsize{8}{9}\selectfont Benchmark Analysis and Foundations} &
\textcolor{black!58}{\textbf{Deciding reason}}\enspace Its tests are defined by a capability query over an expanding pool, and its main result is sparse ordinal aggregation. No fixed full evaluation serves as the reference.\par\smallskip
\textcolor{black!58}{\textbf{Secondary attributes}}\enspace fixed per query, joint pool, item content and historical results, no operational saved unit, ranking \\
\bottomrule
\end{tabularx}
\arrayrulecolor{black}
\caption{\textbf{Examples at category boundaries.} Each paper keeps one primary label. The final column gives the deciding reason, followed by plan timing, benchmark scope, main information, saved unit, and returned result.}
\label{tab:catalogue-boundaries}
\end{table*}

\section{Comparison with Prior Surveys}
\label{app:surveys}

To assess whether this field had been surveyed previously, we supplemented the literature search in Appendix~\ref{app:collection} with survey-focused queries. We combined survey and review with terms related to efficient evaluation, benchmark selection, psychometrics, adaptive testing, benchmark quality, and automated judging, and traced backward and forward citations. By the August 9, 2026 cutoff, we found surveys of related areas, but none that organizes LLM benchmark compression as a unified field covering compression within individual benchmarks, across benchmark suites, and during adaptive evaluation, together with work on redundancy and reliability. To the best of our knowledge, this is the first systematic survey to synthesize this rapidly growing field within a common framework. Our taxonomy and cross-family comparison clarify how the methods relate, what information and costs they require, and how their reliability should be assessed. Prior surveys cover only parts of this scope, as shown in Tables~\ref{tab:adjacent-surveys} and~\ref{tab:broad-surveys}.

\begin{table*}[t]
\centering
\footnotesize
\setlength{\tabcolsep}{2.6pt}
\renewcommand{\arraystretch}{1.18}
\renewcommand{\tabularxcolumn}[1]{m{#1}}
\newcommand{\surveycolhead}[1]{{\fontsize{8}{9.5}\selectfont\bfseries #1}}
\begin{tabularx}{\textwidth}{@{}>{\raggedright\arraybackslash}m{3.05cm}>{\raggedright\arraybackslash}m{2.00cm}*{7}{>{\centering\arraybackslash}X}@{}}
\toprule
\surveycolhead{Survey} & \surveycolhead{Primary focus} & \surveycolhead{Models} & \surveycolhead{Design} & \surveycolhead{Fixed\newline reduction} & \surveycolhead{Adaptive\newline testing} & \surveycolhead{Joint\newline use} & \surveycolhead{Cost} & \surveycolhead{Reliability} \\
\midrule
\textbf{LLM Psychometrics}\newline{\fontsize{8}{9.5}\selectfont\citep{ye2025large}} & Measurement\newline and validation & \yes & \yes & \opt & \opt & \no & \no & \yes \\
\addlinespace[4pt]
\textbf{Computerized}\newline\textbf{Adaptive Testing}\newline{\fontsize{8}{9.5}\selectfont\citep{zhuang2026survey}} & Adaptive tests\newline and control & \yes & \yes & \no & \yes & \no & \yes & \yes \\
\addlinespace[4pt]
\textbf{LLM-as-a-Judge}\newline{\fontsize{8}{9.5}\selectfont\citep{gu2026survey}} & LLM judging\newline and reliability & \no & \opt & \no & \no & \no & \opt & \yes \\
\addlinespace[4pt]
\rowcolor{black!4}\textbf{This survey} & \textbf{Benchmark\newline compression} & \opt & \opt & \yes & \yes & \yes & \yes & \yes \\
\bottomrule
\end{tabularx}
\caption{\textbf{Comparison with adjacent methodological surveys.} The dimensions use general methodological definitions. Models denotes latent-trait, response, calibration, and diagnostic measurement models. Design covers instrument, item-bank, and protocol construction. Fixed reduction selects fewer items or labels before target outcomes are observed. Adaptive testing lets observed outcomes guide acquisition or stopping. Joint use shares evidence or allocation across benchmarks. Cost explicitly considers evaluation resources. Reliability covers uncertainty, calibration, bias, robustness, guarantees, validity, and reference quality. A check marks systematic treatment, Opt. marks explicit but partial treatment, and a cross marks no systematic treatment. Overall, our survey provides the broadest coverage of benchmark compression.}
\label{tab:adjacent-surveys}
\end{table*}

\begin{table*}[t]
\centering
\footnotesize
\setlength{\tabcolsep}{2.6pt}
\renewcommand{\arraystretch}{1.18}
\renewcommand{\tabularxcolumn}[1]{m{#1}}
\newcommand{\surveycolhead}[1]{{\fontsize{8}{9.5}\selectfont\bfseries #1}}
\begin{tabularx}{\textwidth}{@{}>{\raggedright\arraybackslash}m{3.05cm}>{\raggedright\arraybackslash}m{2.00cm}*{7}{>{\centering\arraybackslash}X}@{}}
\toprule
\surveycolhead{Survey} & \surveycolhead{Primary focus} & \surveycolhead{Coverage} & \surveycolhead{Protocols} & \surveycolhead{Evaluators} & \surveycolhead{Quality} & \surveycolhead{Adaptive} & \surveycolhead{Efficiency} & \surveycolhead{Reliability} \\
\midrule
\textbf{LLM Evaluation}\newline{\fontsize{8}{9.5}\selectfont\citep{chang2024survey}} & Evaluation\newline landscape & \yes & \yes & \yes & \opt & \opt & \opt & \opt \\
\addlinespace[4pt]
\textbf{LLM Benchmarks}\newline{\fontsize{8}{9.5}\selectfont\citep{ni2025survey}} & Capabilities\newline and limitations & \yes & \opt & \opt & \yes & \no & \no & \opt \\
\addlinespace[4pt]
\textbf{Static-to-Dynamic}\newline\textbf{Evaluation}\newline{\fontsize{8}{9.5}\selectfont\citep{chen2025benchmarking}} & Contamination\newline resistance & \opt & \opt & \no & \yes & \no & \no & \opt \\
\addlinespace[4pt]
\textbf{Beyond the Leaderboard}\newline{\fontsize{8}{9.5}\selectfont\citep{sheikhi2026beyond}} & Paradigms\newline and lifecycle & \yes & \yes & \yes & \yes & \opt & \opt & \yes \\
\addlinespace[4pt]
\rowcolor{black!4}\textbf{This survey} & \textbf{Benchmark\newline compression} & \opt & \yes & \opt & \yes & \yes & \yes & \yes \\
\bottomrule
\end{tabularx}
\caption{\textbf{Positioning relative to broader LLM evaluation and benchmark surveys.} Coverage includes capability or application domains and benchmark instruments. Protocols covers execution, scoring, and reporting. Evaluators includes human assessment, automatic metrics, and LLM-based judging. Quality covers contamination, saturation, validity, robustness, reproducibility, and lifecycle concerns. Adaptive requires current-target outcomes to guide later acquisition or stopping. Efficiency covers methods designed primarily to reduce evaluation resources. Reliability includes uncertainty, calibration, stability, and formal error control. Symbols follow Table~\ref{tab:adjacent-surveys}. Overall, our survey complements their broad coverage with a focused and systematic treatment of efficient and reliable LLM evaluation.}
\label{tab:broad-surveys}
\end{table*}

\subsection{Adjacent Methodological Surveys}
\label{app:surveys-method}

The closest methodological surveys represent three related traditions. LLM Psychometrics covers construct measurement, instrument design, and validation for LLMs \citep{ye2025large}. The survey of computerized adaptive testing reviews response models, item selection, test control, and stopping beyond the LLM setting \citep{zhuang2026survey}. The LLM-as-a-Judge survey focuses on scalable model-based judging and its reliability \citep{gu2026survey}. Table~\ref{tab:adjacent-surveys} compares them using general methodological dimensions independent of our taxonomy. These surveys cover only parts of our scope, whereas our survey connects fixed test reduction, adaptive testing, joint benchmark use, cost analysis, and reliability in a unified framework for LLM evaluation.

\subsection{Broader Evaluation Surveys}
\label{app:surveys-broad}

Broader surveys map the LLM evaluation landscape rather than focus on cost reduction. Chang et al. organize tasks, benchmarks, and evaluation protocols \citep{chang2024survey}, while Laskar et al. review evaluation methods and practical challenges such as computational demands, dataset subsampling, and human evaluation costs \citep{laskar2024systematic}. Ni et al. catalogue 283 benchmarks and analyze their design and limitations \citep{ni2025survey}, whereas Chen et al. examine contamination-resistant static and dynamic benchmark design \citep{chen2025benchmarking}. Sheikhi et al. synthesize evaluation paradigms, benchmark lifecycles, and methodological weaknesses \citep{sheikhi2026beyond}. Ibrahimoglu et al. present a comprehensive taxonomy and discuss lightweight benchmark subsets, adaptive evaluation, early stopping, and cost analysis \citep{ibrahimouglu2026systematic}. Table~\ref{tab:broad-surveys} recognizes these contributions while distinguishing broad coverage from systematic treatment of evaluation efficiency. We count a dynamic or living benchmark as adaptive only when outcomes from the current target guide later acquisition or stopping. Our survey places less emphasis on cataloguing tasks and benchmarks and instead examines methods that reduce evaluation cost by testing fewer items, benchmarks, or human judgments while preserving reliable scores, rankings, or model-selection decisions.

\section{Taxonomy Guide}
\label{app:taxonomy}

The taxonomy provides a consistent framework for reading, organizing, and counting studies. Its levels address a sequence of questions, and each paper receives one primary label. Other relevant properties are recorded as secondary attributes.

\begin{table*}[t]
\centering
\footnotesize
\setlength{\tabcolsep}{5pt}
\renewcommand{\arraystretch}{1.34}
\begin{tabularx}{\textwidth}{@{}>{\raggedright\arraybackslash}p{3.25cm}>{\raggedright\arraybackslash}p{7.35cm}Y@{}}
\toprule
\rowcolor{white}
\textcolor{black!88}{\textbf{Leaf category}} &
\textcolor{black!88}{\textbf{Use this label when}} &
\textcolor{black!88}{\textbf{Use another label when}} \\
\midrule
\addlinespace[4pt]
\rowcolor{taxsingle!11}
\multicolumn{3}{@{}l@{}}{\hspace{5pt}\taxgroup{taxsingle}{Static single-benchmark compression}} \\
\addlinespace[4pt]
\taxleaflabel{taxsingle}{Content-Based Compression} & The evaluated items are fixed in advance using item or dataset features, or target-independent sampling such as random, stratified, or unequal-probability sampling. The method uses neither past model responses nor a cheap signal from the model being tested. & Past model responses or a signal from the tested model determines which items are evaluated. \\
\taxrowsep
\taxleaflabel{taxsingle}{Historical-Response Compression} & A reusable subset or estimator for one benchmark is built mainly from earlier model responses, score matrices, leaderboard records, or calibrated measurement parameters. & A signal from the model being tested, rather than past responses, determines the evaluated items. \\
\taxrowsep
\taxleaflabel{taxsingle}{Target-Conditioned Compression} & A cheap signal or small pilot from the model being tested selects one fixed batch. Once the batch is fixed, later evaluation results cannot change the selected items or budget. & Evaluation results can change the next item, the remaining budget, or when evaluation ends. \\
\addlinespace[8pt]
\rowcolor{taxcross!10}
\multicolumn{3}{@{}l@{}}{\hspace{5pt}\taxgroup{taxcross}{Cross-benchmark compression}} \\
\addlinespace[4pt]
\taxleaflabel{taxcross}{Cross-Benchmark Compression} & One fixed method jointly reduces work across at least two benchmarks, datasets, or tasks. It returns separate scores or rankings, or a clearly defined suite-level result. & The method is run independently on each benchmark, or its required source runs cost about as much as the evaluations they replace. \\
\addlinespace[8pt]
\rowcolor{taxadaptive!15}
\multicolumn{3}{@{}l@{}}{\hspace{5pt}\taxgroup{taxadaptive}{Adaptive evaluation}} \\
\addlinespace[4pt]
\taxleaflabel{taxadaptive}{Adaptive Item Selection} & Results observed during evaluation determine which item is tested next or how the remaining budget is allocated. The main contribution is the item-selection or budget-allocation strategy. & The main contribution is a rule for stopping or making the final decision. \\
\taxrowsep
\taxleaflabel{taxadaptive}{Sequential Decision and Stopping} & Interim results determine whether to continue, remove a candidate, accept a result, compare models, select a model, or stop. The method is designed mainly to make the final action reliable. & The method mainly selects informative items to estimate a score, with stopping only a secondary component. \\
\addlinespace[8pt]
\rowcolor{taxfoundation!10}
\multicolumn{3}{@{}l@{}}{\hspace{5pt}\taxgroup{taxfoundation}{Analysis and foundations}} \\
\addlinespace[4pt]
\taxleaflabel{taxfoundation}{Benchmark Analysis and Foundations} & The main contribution studies benchmark redundancy, compressibility, measurement, reliability, validity, or the behavior of an existing compression method. It need not introduce a new method. & The main contribution is a tested method that reduces evaluation cost, or compression appears only as background. \\
\bottomrule
\end{tabularx}
\caption{\textbf{Rules for assigning the seven leaf categories.} Apply the scope requirements in Section~\ref{app:taxonomy-scope} first. The middle column gives the defining rule, and the right column identifies the closest competing label.}
\label{tab:taxonomy-guide}
\end{table*}

\subsection{Scope}
\label{app:taxonomy-scope}

This subsection defines which studies enter the taxonomy and how they are labeled. We consider a method benchmark compression if it starts from a predefined full evaluation and evaluates fewer costly units, such as items, benchmark runs, human reviews, judge calls, comparisons, or agent rollouts. It may select all units in advance or choose them adaptively using observed results. From these units, the method must estimate the full-evaluation score or ranking, quantify uncertainty in that estimate, or reproduce the full evaluation's decision. Reducing evaluation cost must be a central goal. Benchmark Analysis and Foundations includes studies that examine benchmark redundancy or compressibility, or assess the accuracy, reliability, and validity of compressed evaluations. We exclude broader work on efficient inference or evaluation unless it meets these criteria. We assign each paper a primary label based on the main problem it addresses, as reflected in its research question, stated goal, formal guarantees, and central experiments. Method components, titles, names, and application settings do not determine the label. A new short benchmark qualifies as compression only if the study explicitly tests its agreement with the predefined full evaluation. If the method changes what the benchmark measures or seeks to improve the benchmark while allowing scores or rankings to change, we classify it as benchmark repair or related analysis.

\subsection{Category Rules}
\label{app:taxonomy-definitions}

Table~\ref{tab:taxonomy-guide} defines each primary label and its nearest boundary. Within Single-Benchmark Compression, the label depends on the information required by the core method. Mentioning or optionally using other information does not change the label.

\begin{table}[t]
\centering
\footnotesize
\setlength{\tabcolsep}{5.5pt}
\renewcommand{\arraystretch}{1.24}
\begin{tabularx}{\columnwidth}{@{}>{\raggedright\arraybackslash\bfseries}p{1.70cm}Y@{}}
\toprule
\rowcolor{appheader}
Dimension & \textbf{Values} \\
\midrule
\rowcolor{appstripe}
Timing & Fixed / one target-conditioned batch\newline Adaptive \\
\addlinespace[2pt]
Scope & Single benchmark / joint suite\newline Source-to-target \\
\addlinespace[2pt]
\rowcolor{appstripe}
Information & Item content / historical results\newline Target-model signal / online results \\
\addlinespace[2pt]
Saved unit & Items / benchmark runs / rollouts\newline Human reviews / judge calls / comparisons \\
\addlinespace[2pt]
\rowcolor{appstripe}
Output & Score / ranking / decision / interval \\
\bottomrule
\end{tabularx}
\caption{Secondary attributes. Multiple values may apply to each dimension.}
\label{tab:taxonomy-attributes}
\end{table}

\begin{table*}[t]
\centering
\footnotesize
\setlength{\tabcolsep}{6pt}
\renewcommand{\arraystretch}{1.18}
\begin{tabularx}{\textwidth}{@{}>{\raggedright\arraybackslash\bfseries}p{2.15cm}YY@{}}
\toprule
\rowcolor{appheader}
Boundary & \textbf{First category} & \textbf{Second category} \\
\midrule
Adaptivity &
\textbf{Target-Conditioned Compression}\newline A limited probe selects one fixed main batch. &
\textbf{Adaptive Evaluation}\newline Main-stage outcomes can change later tests, budget, or stopping. \\
\addlinespace[4pt]
\rowcolor{appstripe}
Benchmark scope &
\textbf{Single-Benchmark Compression}\newline Benchmarks are evaluated independently. &
\textbf{Cross-Benchmark Compression}\newline Joint selection, sharing, skipping, or replacement reduces evaluation across a suite. \\
\addlinespace[4pt]
Main objective &
\textbf{Adaptive Item Selection}\newline Choosing the next item is the main contribution. &
\textbf{Sequential Decision and Stopping}\newline The main guarantee concerns stopping or a downstream action. \\
\addlinespace[4pt]
\rowcolor{appstripe}
Contribution &
\textbf{Operational method}\newline The main contribution is a validated compressor. &
\textbf{Benchmark Analysis and Foundations}\newline Primarily studies the reliability of compression or benchmarks. \\
\addlinespace[4pt]
Reference result &
\textbf{Benchmark compression}\newline Approximates the predefined full-evaluation result. &
\textbf{Benchmark repair / related analysis}\newline Changes the target construct or the reference result. \\
\bottomrule
\end{tabularx}
\caption{\textbf{Common category boundaries.} For mixed information sources, the indispensable source determines the primary label; other sources are recorded as secondary attributes.}
\label{tab:taxonomy-boundaries}
\end{table*}

\subsection{Decision Path}
\label{app:taxonomy-path}

The decision path applies the rules in a fixed order to prevent secondary mechanisms from determining the primary label. \textbf{Contribution type.} First, we distinguish operational methods that reduce evaluation cost from studies that directly analyze benchmark compression or reliability. Operational methods proceed through the remaining branches. Analytical studies are labeled Benchmark Analysis and Foundations, while work outside both scopes is excluded. \textbf{Adaptive evaluation.} Next, we ask whether results obtained during the current evaluation can alter subsequent tests, budget allocation, or stopping. Methods centered on what to test next are labeled Adaptive Item Selection. Those centered on when to stop or what final action to take are labeled Sequential Decision and Stopping. \textbf{Benchmark scope.} For non-adaptive methods, we ask whether a method jointly reduces evaluation across multiple benchmarks or tasks. If so, it is labeled Cross-Benchmark Compression. Applying the same method separately to several benchmarks does not qualify. \textbf{Single-benchmark information.} We classify the remaining methods by the information essential to their operation. Benchmark content yields Content-Based Compression, past model results yield Historical-Response Compression, and a low-cost signal from the current model yields Target-Conditioned Compression.

This ordering reflects the structure of the evaluation process. Adaptivity comes first because current results can change both the test sequence and the stopping point. Among non-adaptive methods, methods that jointly reduce evaluation across multiple benchmarks are separated before single-benchmark methods are classified by their essential information source. When a paper combines several mechanisms, its research question, main objective, principal guarantee, and core experiments determine the primary label. Other mechanisms can be described through secondary attributes.

\subsection{Secondary Attributes}
\label{app:taxonomy-attributes}

To complement the primary taxonomy, Table~\ref{tab:taxonomy-attributes} identifies five secondary dimensions for characterizing methods: plan timing, benchmark scope, main information source, saved unit, and returned result. Multiple values may apply, allowing properties shared across method families to be compared beyond the primary label.

\subsection{Boundary Cases}
\label{app:taxonomy-hybrids}

Method properties can overlap, so Table~\ref{tab:taxonomy-boundaries} clarifies the criteria for common boundary cases. Two authors reviewed these cases using the scope criteria and decision path defined above. Ambiguous assignments were resolved through discussion based on each paper's main contribution and evaluation design. Each paper received one primary label, while other mechanisms were captured as secondary attributes.

\section{Detailed Comparison}
\label{app:comparison}

This section extends the comparison in Section~\ref{sec:comparison} by connecting the required result to selection, recovery, information needs, and cost. It then explains how to validate a method for its intended use.

\subsection{Detailed Problem Formulation}
\label{app:comparison-formal}

Compression starts from a fixed reference evaluation: the models, protocol, units $\mathcal U$, and aggregation rule $G$ are specified before selection. Units may be model-item cells, judgments, comparisons, repetitions, or benchmark and agent runs. Using the notation of Section~\ref{sec:problem},
\[
  \widehat{\theta}_{\mathrm{full}}=G(Y_{\mathcal U}),
  \qquad O_{\mathrm{comp}}=R(Y_{\mathcal S},Z).
\]
Here, $\mathcal S\subseteq\mathcal U$ indexes units with observed reference outcomes, and $Z$ contains auxiliary information available when used. The output $O_{\mathrm{comp}}$ may be an estimate, ranking, decision, interval, or result allowing abstention. A comparable point result is denoted by $\widehat{\theta}_{\mathrm{comp}}$ in the main text. The reference units and weights remain fixed throughout evaluation. For stochastic generation, judging, or rollouts, distinguish recovery of one complete outcome table from estimation of a statistical target, such as expected accuracy. A full run also has sampling error in the latter case, and an interval covers its declared target rather than necessarily another full run's score.

For comparable point results, the goal is $L(\widehat{\theta}_{\mathrm{comp}},\widehat{\theta}_{\mathrm{full}})\leq\varepsilon$ while $C_{\mathrm{comp}}<C_{\mathrm{full}}$, using the same resource and cost scope. Other outputs need the checks in Section~\ref{app:comparison-choice}. These goals guide comparison rather than impose a common guarantee on every method. A fixed-budget method can improve decisions below the full-evaluation cost without stopping early. Reallocating that budget does not itself reduce it, and difficult cases may still require full evaluation.

The central requirement is to preserve a predefined result with fewer or better allocated costly observations. Preserving rankings need not recover absolute means, and finding a winner need not recover every score. A short form supports the properties actually tested. Changing the measured target is benchmark repair and requires separate validation. Foundations studies examine these issues without necessarily proposing a compressor. Faster inference that still collects all the same reference outcomes remains outside this scope.

\subsection{Selection and Recovery}
\label{app:comparison-anatomy}

Selection determines both what to evaluate and whether to continue. Table~\ref{tab:workflows} separates these controls because adaptive selection need not imply early stopping. PromptEval uses a fixed sparse design, while TailoredBench uses a limited probe to select one main batch \citep{polo2024efficient,yuan2025beyond}. Under the convention in Appendix~\ref{app:taxonomy-hybrids}, the latter is Target-Conditioned Compression. Adaptive evaluation instead lets main-stage outcomes change later tests, allocation, or stopping. UCB-E updates allocation within a fixed budget, whereas CELEUS stops when its interval reaches the requested precision \citep{zhou2025speeding,zhou2026celeus}. Fixing the main batch does not make the full two-stage design statistically nonadaptive or its ordinary mean unbiased. Probe reuse must be stated and its cost counted once. These execution patterns can span a single benchmark or a suite. Batch and Rounds describe dependencies between tests, with parallel execution still possible within a round.

Recovery determines what the selected observations support. Table~\ref{tab:evidence-replacement} compares mechanisms that can be combined within a method. An appropriate random design lets a sample mean estimate the reference mean. Cluster weights on deterministic representatives do not provide the same protection. Probability weighting requires valid sampling probabilities and adequate support, with concentrated weights increasing variance. Predictors can use historical responses or fit current sparse observations, as in PromptEval \citep{polo2024efficient}. Their reliability depends on fit and transfer to the target. Residual correction instead uses observed reference errors to adjust predictions, as in CollabEval \citep{fisch2026collabeval}. Under the method's sampling and estimation conditions, inaccurate predictions can reduce precision and savings without invalidating the corrected estimate. Recovery must retain the unit identities, selection probabilities, and test history it needs. Valid stopping remains a separate requirement.

\newcommand{\costlow}{\cellcolor{taxsingle!4}\textbf{Low}}
\newcommand{\costmedium}{\cellcolor{taxsingle!10}\textbf{Medium}}
\newcommand{\costhigh}{\cellcolor{taxsingle!18}\textbf{High}}
\newcommand{\costvariable}{\cellcolor{taxfoundation!12}\textbf{Variable}}

\begin{table*}[t]
\centering
\small
\setlength{\tabcolsep}{8pt}
\renewcommand{\arraystretch}{1.18}
\begin{tabular}{@{}lcccc@{}}
\toprule
Workflow & Initial signal & Update selection & Outcome-based stop & Execution \\
\midrule
Fixed batch & \no & \no & \no & Batch \\
Signal then batch & \yes & \no & \no & Batch \\
Selection only & \opt & \yes & \no & Rounds \\
Stopping only & \opt & \no & \yes & Rounds \\
Selection + stopping & \opt & \yes & \yes & Rounds \\
\bottomrule
\end{tabular}
\captionsetup{width=0.95\textwidth}
\caption{\textbf{Execution patterns.} Checks and crosses mark the presence or absence of a feature. Opt. denotes an optional initial target signal. Updates and stopping use main-stage outcomes, excluding budget exhaustion.}
\label{tab:workflows}
\end{table*}

\begin{table*}[t]
\centering
\small
\setlength{\tabcolsep}{8pt}
\renewcommand{\arraystretch}{1.18}
\begin{tabular}{@{}lcccc@{}}
\toprule
Recovery & Predictions & Residual correction & Key check & Sensitivity \\
\midrule
Sample mean & \no & \no & Sampling & Variance \\
Coreset aggregation & \no & \no & Coverage & Mismatch \\
Probability weighting & \no & \no & Support & Weights \\
Model prediction & \yes & \no & Fit & Shift \\
Prediction + correction & \yes & \yes & Sampling & Residuals \\
\bottomrule
\end{tabular}
\captionsetup{width=0.98\textwidth}
\caption{\textbf{Recovery mechanisms and their checks.} Checks and crosses mark the components used. Residual correction adjusts predictions using observed errors. The final columns summarize the main check and sensitivity.}
\label{tab:evidence-replacement}
\end{table*}

\subsection{Information Requirements}
\label{app:comparison-evidence}

Information sources and recovery mechanisms answer different questions. Table~\ref{tab:signals} describes the information used by each family, while Table~\ref{tab:evidence-replacement} describes how it contributes to estimation. Auxiliary information may include content, historical responses, source-task results, and cheap target signals, but not unobserved target reference labels. Probe includes cheap proxy ratings as well as small reference-quality pilots. Smarter Sampling, for example, uses judge ratings to guide human review \citep{unell2025smarter}. Full-pool generation with sparse review saves labels, not necessarily generation. Source transfer may use $\mathcal S=\varnothing$ when the target result is validated and source costs are included, as illustrated by PACE and BIS \citep{song2026pace,shi2025importance}. Output summarizes the main purpose rather than every return type, so sequential methods can also return intervals or abstain \citep{zhou2026celeus,jung2025trust}.

\begin{table*}[t]
\centering
\small
\setlength{\tabcolsep}{5pt}
\renewcommand{\arraystretch}{1.36}
\begin{tabular*}{\textwidth}{@{\extracolsep{\fill}}lcccccc l@{}}
\toprule
Family & Item/design & History & Probe & Feedback & Suite & Reuse & Output \\
\midrule
Content-Based & \yes & \no & \no & \no & \no & Subset & Estimate \\
Historical-Response & \opt & \yes & \no & \no & \no & Model & Estimate \\
Target-Conditioned & \opt & \opt & \yes & \no & \no & Method & Estimate \\
Cross-Benchmark & \opt & \opt & \opt & \no & \yes & Suite & Suite result \\
Adaptive Item Selection & \opt & \opt & \opt & \yes & \opt & Policy & Estimate \\
\begin{tabular}[t]{@{}l@{}}Sequential Decision\end{tabular} & \opt & \opt & \opt & \yes & \opt & Rule & Decision \\
\bottomrule
\end{tabular*}
\caption{\textbf{Information used by each family.} A check marks defining information, a cross marks excluded information, and Opt. marks optional information.}
\label{tab:signals}
\end{table*}

\subsection{Lifecycle Cost and Operational Risk}
\label{app:comparison-lifecycle}

Cost savings depend on the whole evaluation workflow. For $K$ target evaluations, the total cost in a common unit is
\[
  C_{\mathrm{comp}}^{(K)}=C_{\mathrm{build}}
  +C_{\mathrm{refresh}}^{(K)}+\sum_{k=1}^{K}C_{\mathrm{run},k}.
\]
Each run includes required signals, reference execution, recovery, audits, and any triggered fallback. Count full-pool scans, source runs, and probes once. State whether costs include historical construction or only new work, and amortize assets over actual reuse. Human time, computation, and money require a stated conversion before they can be added. Report wall-clock latency separately because parallelism and sequential dependencies affect elapsed time.

Table~\ref{tab:cost-matrix} shows where each family places its main burden. Historical-Response and Cross-Benchmark Compression invest in reusable assets, so repeated use must outweigh construction and refresh. Target-Conditioned Compression must offset the cost of its signal, while adaptive methods trade parallelism for more focused testing. Near ties and weak transfer can trigger further evaluation and erase expected savings. The qualitative levels describe typical structure rather than measured averages or a universal ranking. Variable applies to outcome-based stopping, while fixed-budget decision methods keep their prescribed budget. Thus, fewer calls alone do not establish lower total cost or latency.

\begin{table*}[t]
\centering
\footnotesize
\setlength{\tabcolsep}{2.5pt}
\renewcommand{\arraystretch}{1.16}
\newcommand{\costhead}[1]{{\fontsize{8}{9.5}\selectfont\bfseries #1}}
\newcommand{\costtext}[1]{\parbox[c][24pt][c]{\linewidth}{\raggedright #1}}
\newcommand{\costfamily}[1]{\costtext{\textbf{#1}}}
\newcommand{\costlevel}[1]{\parbox[c][24pt][c]{\linewidth}{\centering\textbf{#1}}}
\renewcommand{\costlow}{\cellcolor{taxsingle!4}\costlevel{Low}}
\renewcommand{\costmedium}{\cellcolor{taxsingle!10}\costlevel{Medium}}
\renewcommand{\costhigh}{\cellcolor{taxsingle!18}\costlevel{High}}
\renewcommand{\costvariable}{\cellcolor{taxfoundation!12}\costlevel{Variable}}
\begin{tabularx}{\textwidth}{@{}>{\raggedright\arraybackslash}p{3.00cm}*{4}{>{\centering\arraybackslash}p{1.25cm}}>{\raggedright\arraybackslash}p{2.15cm}>{\raggedright\arraybackslash}p{2.20cm}Y@{}}
\toprule
\costhead{Family} & \costhead{Build} & \costhead{Per target} & \costhead{Recovery} & \costhead{Refresh} & \costhead{Reused asset} & \costhead{Main risk} & \costhead{Fallback} \\
\midrule
\costfamily{Content-Based} & \costmedium & \costlow & \costlow & \costlow & \costtext{Fixed plan} & \costtext{Item mismatch} & \costtext{Stratified or\newline full evaluation} \\
\costfamily{Historical-Response} & \costhigh & \costlow & \costmedium & \costhigh & \costtext{Response model} & \costtext{Drift or weak\newline support} & \costtext{Rebuild or\newline sample directly} \\
\costfamily{Target-Conditioned} & \costlow & \costmedium & \costmedium & \costlow & \costtext{Procedure} & \costtext{Weak target\newline signal} & \costtext{Add random\newline reference data} \\
\costfamily{Cross-Benchmark} & \costhigh & \costlow & \costmedium & \costhigh & \costtext{Suite model} & \costtext{Suite or source\newline drift} & \costtext{Keep unsupported\newline parts} \\
\costfamily{Adaptive Item\newline Selection} & \costmedium & \costmedium & \costhigh & \costmedium & \costtext{Policy} & \costtext{Selection bias} & \costtext{Fixed design or\newline full run} \\
\costfamily{Sequential Decision\newline and Stopping} & \costmedium & \costvariable & \costmedium & \costmedium & \costtext{Decision rule} & \costtext{Near ties or\newline invalid stopping} & \costtext{Continue, tie,\newline or abstain} \\
\bottomrule
\end{tabularx}
\caption{\textbf{Typical lifecycle cost and operational risk.} Build covers construction and calibration. Per target covers mandatory signals and selected execution. Recovery covers estimation or decision logic. Refresh covers drift checks and asset updates. The levels summarize common structural demands across each family. Human time, accelerator time, latency, and money remain separate unless a study states a conversion.}
\label{tab:cost-matrix}
\end{table*}
\subsection{Validation Scope}
\label{app:comparison-validation}

Validation should match the transfer required at deployment. Model holdout tests unseen models within a known benchmark or suite. Family holdout withholds entire model families, while time holdout evaluates later models or versions using earlier assets. Task holdout tests tasks excluded from compressor construction. These splits can be combined, and each must cover subset construction, fitting, tuning, and calibration. Model families should be explicitly defined. Test-time probes and unlabeled task descriptions remain available when the method allows them. A time split supports transfer across the tested period.

The split is distinct from how evaluation is performed. Replay and live execution can use the same holdout, while hiding cells for known targets tests recovery within an existing asset. Adaptive replay must reveal only results already acquired along the selected path. Prospective testing freezes the procedure before later targets arrive. Theoretical guarantees provide protection for their stated target, randomness, and assumptions. Neither prospective testing nor a guarantee automatically establishes transfer across all settings.

\subsection{Method Choice and Result Checks}
\label{app:comparison-choice}

The required result determines how compression should be checked. For scores, report estimation error, such as mean absolute error, together with bias. For rankings, examine both overall correlation and reversals among close models. Decision error should match the claimed winner or threshold, with near ties checked separately. Intervals need target coverage and useful width. Selective decisions require conditional risk among decided cases and the decision rate, since abstention is not a correct decision. Distributional estimates need a specified distance and checks of tail error. Quantile estimates need error measured at the stated quantile level. These checks assess the claimed result rather than mean-score accuracy alone.

Method choice follows from these requirements. Match the desired result to the available information in Table~\ref{tab:signals}, the costs and fallbacks in Table~\ref{tab:cost-matrix}, and the deployment setting in Section~\ref{app:comparison-validation}. For fair comparison, align the reference protocol, output, resource, and budget, and include matched random or stratified baselines. Compare error at the same budget or cost at the same quality target. Report variation across random runs and how errors are combined across models and slices. Tail errors and near ties matter when they can change the final decision.

\section{Detailed Analysis}
\label{app:analysis}

This appendix uses representative studies to show how each category selects tests, recovers the target result, and handles failure. The examples are chosen for clear category fit and methodological diversity rather than exhaustive coverage. For each operational category, we identify the saved cost, key assumptions, validation target, failure conditions, and fallback. The first six categories compress evaluation, while Benchmark Analysis and Foundations assesses the reliability of both compression methods and their underlying benchmarks. Appendix~\ref{app:catalogue} provides the complete catalogue, and Appendix~\ref{app:comparison} presents a cross-category comparison.

\subsection{Content-Based Compression}
\label{app:analysis-content}

Content-Based Compression builds a fixed subset from a sampling rule or item information before the target model is evaluated. The full result may come from direct averaging, design weights, or a fitted recovery model.

\noindent\textbf{Coverage and correction.}
This category includes random and stratified sampling, structured sparse designs, and item-data coresets. In fixed designs, auxiliary scores can instead set sampling probabilities or calibrate automated judgments \citep{purificato2026select,divekar2026statistically}. PromptEval spreads a fixed budget across a sparse prompt-by-example matrix and fits an item response model to recover prompt performance distributions \citep{polo2024efficient}. Scales++ groups items by cognitive-demand descriptions and combines group weights with predictions for untested items \citep{bean2025scales++}. These routes save model-item runs or human labels. Validation compares the compressed score, distribution, or judge measure with its full-pool counterpart.

\noindent\textbf{When it can fail.}
PromptEval relies on the validity of its response model, while Scales++ relies on its cognitive descriptions and regression fit \citep{polo2024efficient,bean2025scales++}. Probability-based sampling and judge correction likewise require accurate sampling information or a matched gold subset \citep{purificato2026select,divekar2026statistically}. Beyond these method-specific assumptions, small subsets may have high variance, and item features may not capture latent skills. If repeated samples are unstable or less accurate than a random baseline, broaden the stratification or revert to the full benchmark.

\noindent\textbf{Takeaway and boundary.}
The original benchmark result can be recovered with a sample mean, a design-weighted estimator, or a fitted model. Reweighting alone does not create a new benchmark. The method becomes benchmark repair only when it changes the target score or the construct being measured.

\subsection{Historical-Response Compression}
\label{app:analysis-history}

Historical-Response Compression reuses past evaluation results to choose tests or recover results for untested items. These results form a model-by-item response matrix. Its reliability depends on whether earlier models cover the behavior of the new one.

\noindent\textbf{Reuse earlier results.}
Anchor Points groups items by source-model confidence and uses weighted medoids to estimate the full score \citep{vivek2024anchor}. tinyBenchmarks fits a multidimensional item response model and selects anchors that cover the resulting item parameters \citep{polo2024tinybenchmarks}. SparseEval learns anchors and weights from a historical model-item matrix to predict full scores \citep{zhang2026sparseeval}. CollabEval uniformly samples target cells, fills the remaining matrix, and corrects the estimates using errors observed on tested cells \citep{fisch2026collabeval}. All four methods save target model-item runs. Their validation should hold out complete target models and compare the recovered scores, differences, or intervals with the corresponding full results.

\noindent\textbf{When it can fail.}
Anchor Points, tinyBenchmarks, and SparseEval depend on historical patterns transferring to the new target \citep{vivek2024anchor,polo2024tinybenchmarks,zhang2026sparseeval, huang2026zipbenchlowcostframeworkcompressing}. CollabEval also requires random sampling and stable matrix completion \citep{fisch2026collabeval}. More broadly, a small, stale, or narrow model collection can misrepresent a new model family. If held-out target error is unstable or exceeds the error of a random subset, add random target tests, refresh the matrix, or run the full benchmark.

\noindent\textbf{Takeaway and boundary.}
Earlier scores can guide subset selection or recover skipped results, but they can also pull a new model toward outdated patterns. A method moves into Target-Conditioned Compression if the current model determines the source models or the fixed test set.

\subsection{Target-Conditioned Compression}
\label{app:analysis-target}

Target-Conditioned Compression uses a low-cost signal from the current model to define a target-specific batch before collecting expensive test outcomes or labels. This signal may come from a small probe, a full-pool output scan, confidence scores, or inexpensive proxy ratings.

\noindent\textbf{Use a cheap target signal.}
TailoredBench runs a small shared set, finds source models with similar answers, and expands that set into target-specific medoids. Results from the source models then recover the target score and ranking \citep{yuan2025beyond}. MAD scans candidate outputs for large semantic differences and asks humans to compare only selected pairs \citep{feng2025sample}. Smarter Sampling uses cheap judge ratings to choose a smaller human audit set and estimate full-pool judge reliability \citep{unell2025smarter}. TailoredBench saves model-item runs, while MAD and Smarter Sampling save human reviews. Validation targets score or rank recovery for the former, ranking for MAD, and judge reliability for Smarter Sampling.

\noindent\textbf{When it can fail.}
TailoredBench reports that probe error can affect both source matching and recovery \citep{yuan2025beyond}. MAD pays for full-pool generation and discrepancy search, while Smarter Sampling relies on small-sample approximations \citep{feng2025sample,unell2025smarter}. More broadly, a cheap current-model signal may not match high-quality labels. If random high-quality checks change the ranking or reliability estimate, expand the audit or run the full evaluation.

\noindent\textbf{Takeaway and boundary.}
The signal scan, candidate generation, cheap scoring, and high-quality checks all count toward total cost. This category fixes the expensive batch after the low-cost signal is collected. It becomes adaptive only when each new high-quality result can change what is tested next or when testing stops.

\subsection{Cross-Benchmark Compression}
\label{app:analysis-cross}

Cross-Benchmark Compression reduces suite-wide evaluation cost through a shared selection or transfer rule. It may sample items across benchmarks, omit entire tasks, or estimate performance on an expensive target from a cheaper source.

\noindent\textbf{Share work across benchmarks.}
metabench chooses informative item-response anchors from several benchmarks. Target results on these anchors recover each benchmark score, a suite average, and local intervals \citep{kipnis2025metabench}. SimBA instead treats a whole benchmark or task as the selection unit. It predicts skipped scores and ranks from tested columns in a historical performance matrix \citep{subramani2025simba}. PACE uses cheap source items to predict an expensive agent benchmark across calibration models \citep{song2026pace}. BIS reweights cached source-benchmark results to estimate the target score \citep{shi2025importance}. Validation must use new models or benchmarks that were not used to learn the shared relation.

\noindent\textbf{When it can fail.}
metabench needs a broad historical suite, and SimBA assumes that benchmark relations persist \citep{kipnis2025metabench,subramani2025simba}. PACE can become unstable with a small calibration fleet, while BIS suffers when source and target distributions overlap poorly \citep{song2026pace,shi2025importance}. More broadly, transfer saves little when the source is neither cached nor cheaper. If error rises on held-out models or benchmarks, run the unsupported target directly.

\noindent\textbf{Takeaway and boundary.}
These routes save different units, including item runs, whole tasks, and expensive target runs. Cost comparisons must use the same saved unit and the same final result. Compressing several benchmarks separately still belongs to Single-Benchmark Compression.

\subsection{Adaptive Item Selection}
\label{app:analysis-acquisition}

Adaptive Item Selection uses each new target result to choose the next test. The final score must account for how those tests were chosen.

\noindent\textbf{Select, update, and correct.}
Fluid updates an item response ability estimate after every answer and chooses the unused item with the most information. The final estimate gives the score and ranking \citep{hofmann2025fluid}. Active Evaluation Acquisition uses tested scores, prompt embeddings, and neural-process predictions to choose each next prompt. It recovers scores or win rates from tested and predicted results \citep{li2024active}. FAQ-PAI narrows the score interval while retaining uniform exploration. It corrects factor predictions with direct tests and recorded sampling probabilities \citep{wu2026efficient}. All three save target-item tests. Validation compares their recovered result with a full run and repeats the acquisition path.

\noindent\textbf{When it can fail.}
Fluid uses one binary ability axis, which can miss skill-specific behavior \citep{hofmann2025fluid}. Active Evaluation Acquisition needs many complete training evaluations, and FAQ-PAI relies on variance and sampling-probability conditions \citep{li2024active,wu2026efficient}. More broadly, adaptive estimates are unstable when a few items receive very large weights. If repeated runs produce different scores, recalibrate, increase uniform exploration, or run the full benchmark.

\noindent\textbf{Takeaway and boundary.}
Choosing informative items does not by itself make the final score reliable. Researchers must record every selection, check coverage, and state how the final score is calculated. The category changes to Sequential Decision and Stopping when the main result is a final choice or a stopping guarantee.

\subsection{Sequential Decision and Stopping}
\label{app:analysis-stopping}

Sequential Decision and Stopping targets the final action. Fixed-budget methods seek the best decision within a prescribed cost, while fixed-confidence methods continue until they meet a specified accuracy or agreement threshold.

\noindent\textbf{Stop, choose, or escalate.}
Flash-HELM tests increasingly larger uniform samples and stops when a leaderboard tier is clear enough \citep{perlitz2024efficient}. UCB-E assigns more tests to candidates with optimistic scores and returns the best model within a fixed budget \citep{zhou2025speeding}. It belongs here because allocation serves the final model choice rather than every full score. CELEUS corrects predicted scores with direct tests and stops when its interval reaches the requested width \citep{zhou2026celeus}. Trust or Escalate routes uncertain pairwise judgments to a stronger judge or a human and abstains when no judge qualifies \citep{jung2025trust}. These outputs are choices, intervals, or abstentions. Validation must test the stated decision rule rather than only rank correlation.

\noindent\textbf{When it can fail.}
Flash-HELM's tier rules are tied to its leaderboard, and UCB-E assumes a fixed score matrix \citep{perlitz2024efficient,zhou2025speeding}. CELEUS requires bounded scores and predictable sampling choices, while Trust or Escalate needs calibration to transfer to the test pool \citep{zhou2026celeus,jung2025trust}. Near-tied candidates may still use almost the full budget. If the stopping or agreement rule is not met, continue testing, abstain, or request human review.

\noindent\textbf{Takeaway and boundary.}
This category covers allocation for final selection, elimination, escalation, abstention, or stopping. A method belongs to Adaptive Item Selection instead when its main aim is choosing the next item for a score estimate.

\subsection{Benchmark Analysis and Foundations}
\label{app:analysis-foundations}

Benchmark Analysis and Foundations asks whether a benchmark can be shortened without losing a meaningful comparison. It also checks whether the scoring model and the benchmark itself are reliable.

\noindent\textbf{Check before compressing.}
BenchScope estimates the effective dimension of a historical response matrix and keeps tasks that add new information. It then compares the reduced ranking with the full ranking to measure redundancy \citep{sha2026benchscope}. Can We Trust IRT? tests item response estimators when few models are available or model abilities cover only a narrow range. It shows how unstable item parameters can distort rankings from short forms \citep{jiang2026can}. Resolution Diagnostics uses paired score gaps, correct rates, and correlation to calculate the sample size needed for a stated difference and error level \citep{kotawala2026resolution}. If the sample is too small, the comparison should remain unresolved. BENCHMARK2 keeps items with stronger capability alignment and discrimination, then scores models on the revised set \citep{qian2026benchmark}. Because this changes what the benchmark measures, the revised benchmark requires separate validation.

\noindent\textbf{When it can fail.}
BenchScope depends on the current model population, and Can We Trust IRT? does not settle multidimensional settings \citep{sha2026benchscope,jiang2026can}. Resolution Diagnostics assumes that prompts represent the intended population, while BENCHMARK2 can rely on circular peer comparisons \citep{kotawala2026resolution,qian2026benchmark, ling2026neuralchainofthoughtsearchsearching}. More broadly, few models or a narrow ability range can make a short form unstable. If rankings change across model groups or small subset changes, collect more results, keep the full benchmark, or repair it first.

\noindent\textbf{Takeaway and boundary.}
These studies guide the six operational categories rather than form another compression process. They identify redundancy, unreliable scoring, unresolved comparisons, and benchmarks that need repair. A study that runs a smaller test set and recovers the original result belongs to an operational category. A study that diagnoses or redesigns the reference benchmark belongs here.

\section{Literature Collection}
\label{app:collection}

To make the literature collection process easy to understand and reproduce, in this section, we describe the search strategy, iterative expansion, screening criteria, and version consolidation.

\paragraph{Search scope and query design.} We set the corpus cutoff to August 9, 2026 and considered published papers and public preprints available by that date. We searched Google Scholar, Semantic Scholar, OpenAlex, arXiv, ACL Anthology, OpenReview, ACM Digital Library, IEEE Xplore, and PMLR. We used Crossref, DBLP, OpenCitations, and official publication pages to verify metadata and citation links. We also conducted venue-focused searches across ACL, EMNLP, NAACL, EACL, AACL, TACL, COLM, ICML, NeurIPS, ICLR, AAAI, IJCAI, KDD, SIGIR, CVPR, ICCV, and ECCV. Initial queries combined an evaluation object term with one or more mechanism or goal terms from the families below. We adapted the syntax to each source while preserving these combinations. These searches produced the seed papers for subsequent expansion.

\begin{center}
\begin{minipage}{0.98\columnwidth}
\small
\setlength{\tabcolsep}{4pt}
\setlength{\parskip}{0pt}
\renewcommand{\arraystretch}{1.16}
\begin{tabularx}{\linewidth}{@{\hspace{4pt}}>{\fontsize{8.5}{10}\selectfont\hyphenpenalty=10000\relax}X@{\hspace{4pt}}}
\toprule
\rowcolor{appheader}
\multicolumn{1}{@{\hspace{4pt}}l@{\hspace{4pt}}}{\textbf{Query families and representative terms}} \\
\midrule
\textbf{Evaluation object.} LLM, language model, benchmark, evaluation suite, leaderboard \\[4pt]
\rowcolor{appstripe}
\textbf{Efficiency and subsets.} efficient evaluation, data-efficient evaluation, sample-efficient evaluation, fewer items, subset selection, coreset selection, short benchmarks, micro benchmarks, sparse benchmarks, partial benchmarks \\[4pt]
\raggedright\arraybackslash\textbf{Prediction and recovery.} score prediction, rank prediction, response-matrix completion, collaborative filtering, \mbox{low-rank modeling} \\[4pt]
\rowcolor{appstripe}
\textbf{Adaptive and measurement.} active testing, adaptive testing, sequential stopping, item response theory, psychometrics, multi-fidelity evaluation, redundancy, reliability \\[4pt]
\bottomrule
\end{tabularx}
\end{minipage}
\end{center}

\paragraph{Iterative expansion.} Starting from the seed papers, we applied the same expansion procedure to every eligible paper. We extracted new terms for further keyword queries, searched its references and indexed forward citations, and followed relevant authors, projects, and venues. After deduplication and screening, each newly included paper triggered another search round. We repeated this cycle until a final sweep at the corpus cutoff found no additional eligible papers, then froze the corpus.

\paragraph{Screening.} We deduplicated records using DOI, arXiv or OpenReview identifiers, URLs, and normalized titles. Two authors then screened titles and abstracts. When eligibility was unclear, they reviewed the relevant method and experiment sections. We included an operational method only if it began with a predefined full evaluation of LLMs, reduced or adaptively allocated executed units or other costly evidence, used the retained evidence to recover a full-evaluation score, ranking, or model-selection decision, and treated cost reduction as a primary contribution. We separately included studies that directly examined LLM benchmark redundancy, measurement, reliability, or compression validity. Mixed-domain papers required a distinct LLM experiment supporting the relevant claim. We excluded work centered on training-data selection, prompt selection, model or context compression, and standard inference acceleration. We also excluded benchmark construction that did not preserve a result from a predefined full evaluation, methods that only reduced per-unit evaluator cost while retaining complete reference evaluation coverage, and studies without direct LLM evidence.

\paragraph{Version consolidation.} We grouped versions using identifiers, normalized titles, authors, years, and abstracts. The formal publication served as the canonical record when available. Preprints, workshop versions, earlier titles, and alternate URLs were retained as aliases. Records were merged only after their relationship was verified.

\section{Additional Statement}
\label{app:additional-statement}

This section summarizes our use of AI assistants and the broader impact of this survey.

\paragraph{Information about AI Use.}
AI assistants were used in this work exclusively for English language polishing and assisting with literature collection and verification. All research ideas, analyses, taxonomy design, and scientific conclusions are original contributions of the authors. The authors reviewed all LLM-assisted outputs and take full responsibility for the final content of this paper.

\paragraph{Broader Impact.}
This survey provides a shared foundation for advancing efficient and reliable LLM evaluation. Its unified taxonomy, comparative analysis, and research agenda help researchers navigate the literature, compare methods on common grounds, and identify promising directions. By clarifying information requirements, lifecycle costs, and reliability, it supports informed method selection and more transparent reporting of evaluation results. This guidance can help practitioners assess where benchmark compression offers meaningful savings while preserving the evidence needed for model comparison. More broadly, by promoting efficient and reproducible model evaluation, the survey can help reduce computational demands and associated financial and environmental costs, contributing to more sustainable AI research.

\section{Survey Contributions}
\label{app:contributions}

The survey is designed to make a fragmented area easy to identify, classify, understand, and use.

\noindent\textbf{First survey of a rapidly emerging field.} Research on benchmark compression for LLM evaluation emerged in 2024 and grew rapidly through 2026. To the best of our knowledge, no prior survey has organized it as a unified field. We bring together methods for compressing individual benchmarks and benchmark suites, adapting evaluation as results arrive, and analyzing redundancy and reliability. Within a unified framework, we compare their information needs, costs, strengths, limitations, and suitable settings. This synthesis guides method selection and validation. Its common formulation makes methods and results more comparable, supporting consistent reporting and future quantitative synthesis. By advancing efficient and reliable LLM evaluation, the survey can also reduce unnecessary computation and energy use, along with their financial and environmental costs.

\noindent\textbf{Intuitive taxonomy with clear decision rules.} The hierarchy provides an intuitive and coherent map for navigating the literature and comparing methods across evaluation settings. Following a compact decision tree based on each paper's primary contribution and evaluation design yields a clear primary label, making category boundaries explicit and classification reproducible. The primary label captures the paper's defining contribution, while secondary attributes enable finer-grained comparison of its methodological design. Appendix~\ref{app:taxonomy-definitions} formalizes the category definitions, Appendix~\ref{app:taxonomy-path} presents the decision tree, and Appendix~\ref{app:taxonomy-hybrids} explains overlaps and boundary cases.

\noindent\textbf{Method roadmap and detailed analysis.} Sections~\ref{sec:static}--\ref{sec:foundations} provide a roadmap of the seven categories and explain in detail how each family works, its strengths and limitations, and the settings in which it is most suitable. Appendix~\ref{app:analysis} further presents representative studies from each category and explains their main ideas and designs in greater depth. Together, these discussions connect the field-level taxonomy to concrete methods, helping readers grasp the literature, identify approaches for their own needs, and find promising directions for further research. By making the field easier to understand and use, this survey lowers the barrier to entry and provides a foundation for developing more efficient and reliable LLM evaluation methods.

\noindent\textbf{In-depth comparison across method families.} Section~\ref{sec:comparison} compares the information each family uses, how it estimates full-evaluation results, its total costs, limitations, and suitable settings. Appendix~\ref{app:comparison} extends this analysis with detailed comparisons of subset selection, score recovery, information requirements, validation methods, failure conditions, and fallbacks. Our lifecycle cost analysis explicitly accounts for constructing, calibrating, and refreshing compression schemes alongside evaluation costs for each target. This perspective helps readers assess whether savings across repeated evaluations justify construction and maintenance costs and choose methods suited to their evaluation scale and resources. Together, these comparisons support realistic efficiency reporting, informed deployment, and the development of more efficient and reliable LLM evaluation methods.

\noindent\textbf{Research agenda for efficient evaluation.} Beyond synthesizing existing methods, the survey identifies how compression research can shape the benchmarks used by the wider community. Section~\ref{sec:challenges} connects generalization across models and over time, updating established benchmarks, and compression before release into an agenda spanning benchmark design, adoption, and maintenance. This agenda treats efficiency as a lasting property of shared benchmarks and aligns cost reduction with continued relevance and reliable model comparison. It gives researchers and benchmark developers common priorities for reducing repeated evaluation costs, widening participation, and sustaining informative evaluation as model capabilities evolve.

\FloatBarrier

\end{document}